# A novel SDASS descriptor for fully encoding the information of 3D local surface

Bao Zhao[a,b], Xinyi Le[b], Juntong Xi[a,b,c,1]

[a] State Key Laboratory of Mechanical System and Vibration, Shanghai Jiao Tong University, Shanghai 200240, China
[b] School of Mechanical Engineering, Shanghai Jiao Tong University, Shanghai, 200240, China
[c] Shanghai Key Laboratory of Advanced Manufacturing Environment, Shanghai Jiao Tong University, Shanghai 200240, China

**ABSTRACT**

Local feature description is a fundamental yet challenging task in 3D computer vision. This paper proposes a novel descriptor, named Statistic of Deviation Angles on Subdivided Space (SDASS), of encoding geometrical and spatial information of local surface on Local Reference Axis (LRA). In terms of encoding geometrical information, considering that surface normals, which are usually used for encoding geometrical information of local surface, are vulnerable to various nuisances (e.g., noise, varying mesh resolutions etc.), we propose a robust geometrical attribute, called Local Minimum Axis (LMA), to replace the normals for generating the geometrical feature in our SDASS descriptor. For encoding spatial information, we use two spatial features for fully encoding the spatial information of a local surface based on LRA which usually presents high overall repeatability than Local Reference Axis (LRF). Besides, an improved LRA is proposed for increasing the robustness of our SDASS to noise and varying mesh resolutions. The performance of the SDASS descriptor is rigorously tested on four popular datasets. The results show that our descriptor has a high descriptiveness and strong robustness, and its performance outperform existing algorithms by a large margin. Finally, the proposed descriptor is applied to 3D registration. The accurate result further confirms the effectiveness of our SDASS method.

## 1. Introduction

Local feature descriptor being used for encoding the information of local surface has many applications in 3D computer vision areas such as 3D registration [1-3], 3D object categorization and recognition [4-6], 3D model retrieval and shape analysis [7-8] and 3D biometric [9], to name a few. In the last few years, with the flourishment of low-cost sensors and high-speed computing systems, 3D data becomes easily available, which further improves the significance of investigating local shape descriptors in 3D computer vision area.

A local shape descriptor is usually constructed by transforming the geometrical and spatial information of a local surface into a feature vector representation [10]. Noting that the descriptor presented in this paper is applied to rigid objects. Therefore, the fundamental attribute of local shape descriptor should be invariant to rigid transformation. Furthermore, a shape descriptor should have a high descriptiveness and strong robustness [11]. The descriptiveness of a local shape descriptor represents an ability of encoding the predominant information on the underlying local surface. In other words, the descriptiveness denotes the ability of distinguishing one local surface from another. The robustness of a local feature descriptor indicates an ability of resisting the impact of various nuisances including noise, varying mesh resolutions, etc. Besides, the compactness and efficiency are also important to a feature descriptor for some time-crucial applications including robots and mobile phone [12]. So, designing a local feature descriptor with comprehensive good performance for dealing with above mentioned nuisances is a tremendous challenge. Over the last two decades, a number of local feature descriptors are designed for improving the ability of coping with these nuisances. Examples include Spin Images (SI) [13-14], Unique Shape Context (USC) [15], Tri-Spin-Image (TriSI) [16], to name a few. For more details, readers can refer to a recent survey [11]. For a local feature descriptor, local frame and feature representation are two major elements for determining its performance. The local frame can be divided into two categories. One is defined as local reference frame (LRF), and another is defined as local reference axis (LRA). The LRF is composed of three orthogonal axes, and the LRA only comprises a single orientated axis. Therefore, LRF can provide entire local 3D spatial information including radial, azimuth and elevation directions, while LRA only provide the spatial information in radial and elevation directions. It is worth noting that some local descriptors (e.g.

[1] Corresponding author.

*E-mail addresses*: zhaobao1988@sjtu.edu.cn (B. Zhao); lexinyi@sjtu.edu.cn (X. Le); jtxi@sjtu.edu.cn (J. Xi).

THRIFT [17], PFH [18]) do not use both the LRF and LRA to construct their features. These descriptors can encode the spatial information only in radial direction, which usually present a limiting descriptiveness owing to the lack of adequate spatial information [11]. For the LRF/A-based descriptors, although LRF can provide entire spatial information, the repeatability of x/y axis are more vulnerable to nuisances (e.g. noise, varying mesh resolution and symmetrical surface) than z axis, and more time is needed for constructing it than LRA [12, 19]. Therefore, a local descriptor constructed on LRA has a high potential with robust to various nuisances [12]. For LRA-based descriptors, the repeatability of LRA directly influence the performance of these descriptors. To achieve a high repeatability of LRA, many methods, such as the techniques proposed in [10, 20], have proposed for constructing LRA or LRF (the z-axis of an LRF can be used as an LRA). In these methods, LRA or the z-axis of LRF is usually constructed by covariance analysis.

In the view of the feature representation of a descriptor, the geometrical and spatial information of a local surface are usually encoded for representing the local shape. Some existing descriptors only encode the geometrical information of a local surface [17, 21]. In these methods, the deviation angle between normals or between normal and LRA is a popular way for encoding local geometrical information. However, owing to low repeatability of normal, geometrical information of local surface encoded by the deviation angle presents a lower robustness [11]. Therefore, these descriptors of only encoding geometrical information often present a poor performance for resisting noise, varying mesh resolutions, etc. In contrast, some descriptors only encode the spatial information of a local surface [13, 16, 22]. However, some of these descriptors are incomplete to encode spatial information by transforming 3D information to 2D/1D information (e.g., TriSI [16]), and some of these descriptors are redundant to encode the spatial information by repetitively using the information of x, y and z coordinates of local points (e.g., RoPS [22], TOLDI [23]). In addition, some descriptors encode both geometrical and spatial information of a local surface [3, 24]. Although, encoding geometrical information together with spatial information can improve the descriptiveness of a feature descriptor [24], as mentioned above, low robustness of geometrical feature and imperfect technique of encoding spatial information also limit the descriptiveness and robustness of these descriptors.

In these regards, we propose a novel local feature descriptor named Statistic of Deviation Angles on Subdivided Space (SDASS). Our SDASS is generated on LRA and encodes both geometrical and spatial information of a local surface. Specifically, we first propose an improved LRA which is developed from the LRF proposed by Yang et.al [23]. Considering normal being vulnerable to various nuisances, in order to improve the robustness of encoding geometrical information, we propose a new geometrical attribute named Local Minimum Axis (LMA), which is verified with a strong robustness to resist various nuisances, to replace normals for constructing the deviation angle between LMA and LRA. For encoding spatial information, our SDASS use two spatial features for fully encoding spatial information of a local surface on LRA. Different from some previous local descriptors which need to preprocess initial point cloud such as triangulation [16, 22], our SDASS is directly generated on initial point clouds. For verifying the performance of the SDASS, it is applied to four popular datasets, and compared to some state-of-the-arts. The experimental results show that the performance of our SDASS exceeds the exiting methods by a large margin. In addition, the SDASS is applied to 3D registration in the last of this paper. The accurate results further confirm the superiority of the SDASS descriptor. The major contributions of this paper are summarized as follows.

*First*, we propose a geometrical attribute LMA which has a significantly high repeatability compared to normals. We use the proposed LMA to replace normals for generating deviation angles, which presents a high descriptiveness and strong robustness for encoding the geometrical information of local surface. *Second*, a novel local shape descriptor named SDASS is proposed. The SDASS is generated on LRA and describes local shape by the combination of encoding geometrical and spatial information. The experimental results show that the performance of SDASS significantly surpasses the existing local shape descriptors. *Third*, an improved LRA is proposed, which has a superior performance for resisting noise and varying mesh resolutions compared to the existing LRF/A.

The remainders of this paper is organized as follows. Section 2 provides a brief literature review for generating LRF/A and local feature descriptors. Section 3 details the process of generating our SDASS method. Section 4 introduces the evaluation results of our SDASS descriptor together with some state-of-the-arts on four popular datasets. Section 5 describes the comparison between our SDASS descriptor and some state-of-the-arts on the application of 3D registration. The conclusions and several future works are drawn in Section 6.

## 2. Related work

This section presents a brief overview of the existing methods for local surface feature description. Considering that the SDASS method belong to the family of LRA-based methods. The existing methods for constructing LRA or the z-axis of LRF are first reviewed. Then, the existing local shape descriptors are divided into three categories to be described respectively.

### *2.1 The methods of constructing LRA or LRF*

Most existing methods of constructing LRA or the z-axis of LRF are based on covariance analysis [19]. In these methods, the z-axis or LRA is the normalized eigenvector corresponding to the minimal eigenvalue of a covariance matrix. Mian et al. [25] used radius neighbors instead of *k* nearest neighbors to construct the covariance matrix for improving the robustness to varying mesh resolutions. However, the sign of the LRF is not defined unambiguously. Tombari et al. [10] constructed a weighted covariance matrix by first using a key point to replace the barycenter of radius neighbors, and then assigning smaller weights to more distant points. The sign of this LRF is unambiguous by inclining the barycenter of local surface. This technique has been verified with a good robustness to noise, while is vulnerable to varying mesh resolutions [23]. Later, Guo et al. [22] presented a novel method for constructing LRF by first employing two weighting strategies to each triangle on a local surface, and then using all weighted triangles to calculate the covariance matrix. The sign of this LRF is disambiguated by aligning the direction to the majority of the point scatter. The main advantage of this method is to have a high robustness to varying mesh resolutions [22], while the efficiency is very low [23]. Recently, Petrelli et al. [1] and Yang et al. [23] extracted a small subset from the radius neighbors for constructing covariance matrix. The main purpose of using less points for calculating the z-axis is to improve its robustness to occlusion, clutter and mesh boundaries. In addition, Johnson et al. [13] and Yang et al. [3] directly use normal as LRA. Since normal is vulnerable to various nuisances (e.g. noise, varying mesh resolution), the LRA present a low robustness.

Some methods mentioned above have nicely solved the problem of sign ambiguous and are robust to some nuisances. However, they do not have an overall good performance for coping with all nuisances (e.g. noise, varying mesh resolution, etc.), such as the LRF presented by Guo et al. [22] has a strong robustness to noise and mesh resolution variation while it is time consuming and vulnerable to mesh boundary, and the LRF presented by Yang et.al. [23] is robust to mesh boundary while is vulnerable to noise and varying mesh resolutions.

### *2.2 Local feature descriptors*

Local feature descriptors have been widely presented in literatures over the last two decades [11, 26]. Among these descriptors, some only encode geometrical information, some only encode spatial information, and others encode the combination of geometrical and spatial information of a local surface.

There are some descriptors of only encoding geometrical information. Flint et al. [17, 27] proposed a local feature descriptor called THRIFT which uses the deviation angles between the normal at a key point and the normals at its neighbors to construct a 1D histogram. The THRIFT is a very efficient descriptor while is very sensitive to noise and varying mesh resolutions [11]. Rusu et al. [18] proposed a point feature histogram (PFH) by using several features of point pairs in a support region. These features are calculated on a Darboux frame constructed by the normals and point positions of the point pairs. The PFH is robust to varying mesh resolutions while is vulnerable to noise [11]. Later on, in order to improve the efficiency, they used the simplified point feature histogram (SPFH) of neighbors to construct the fast point feature histograms (FPFH) descriptor [21] which has a similar performance with PFH in descriptiveness and robustness.

There are some existing descriptors of only encoding spatial information. Johnson [13] presented a local shape descriptor named Spin Image (SI). The Spin Image use the normal at a key point as the LRA, and then each point in the support region is represented by two spatial distances. Finally, the Spin Image is generated by accumulating the number of local points into each bin of the 2D array. The Spin Image completely encodes the spatial information of a local surface on LRA while is sensitive to noise [11] owing to the low repeatability of its LRA. Tombari et al. [15] proposed a Unique Shape Context (USC) which is developed from 3DSC descriptor [28].The USC is generated on an LRF, proposed by Tombari et al. [10], by dividing the local 3D space into bins along azimuth, elevation and radial directions. The USC completely encode the spatial information on LRF and present a high robustness to noise, clutter and occlusion, while is sensitive to mesh resolution variation [11]. Guo et al. [20, 22] presented the Rotational Projection Statistics (RoPS) descriptor. In RoPS descriptor, a novel LRF is first constructed, and then the feature representation is generated by computing a series of statistics of point density with respect to numerous rotations of local surface around each axis. The RoPS descriptor is proved to have a high descriptiveness [11], while is very time-consuming. Like the view-based strategy in RoPS descriptor, Guo et al. [16, 29] also presented a Tri-Spin-Image (TriSI) descriptor. TriSI is also generated on an LRF which is constructed by a similar technique of constructing the LRF in RoPS. Then, TriSI is generated by concatenating three Spin Image signatures created on each axis of the LRF. The performance of TriSI to resist various nuisances (e.g. noise, varying mesh resolutions, etc.) is slightly better than that of the RoPS [11]. Like RoPS, TriSI is also time-consuming. Recently, Yang et al. [23] proposed a novel Triple Orthogonal Local Depth Images (TOLDI) descriptor by first constructing an LRF, and then concatenating three local depth images obtained from three orthogonal view planes in the LRF to generate a feature vector. Although TOLDI achieve a good performance for resisting various nuisances, the feature vector of TOLDI present a low compactness [23].

There are some descriptors are generated by encoding the combination of geometrical and spatial information. Tombari et al. [10, 24] presented the SHOT feature descriptor based on an LRF. The SHOT descriptor first encodes spatial information on the LRF by dividing the spherical neighborhood space into several bins along the radial, azimuth and elevation directions. Then, for each bin, the geometrical information is encoded by generating the deviation angles between the normal at the key point and the normals at the radius neighbors. Despite SHOT having a high descriptiveness, it is sensitive to mesh resolution variation [11, 23]. Recently, Yang et al. [3] proposed a Local Feature Statistics Histograms (LFSH) by concatenating three features, including two spatial features (local depth and horizontal projection distance) and one geometrical feature (deviation angle), to generate a feature vector. The LFSH descriptor is very compact and efficient while suffers from limited descriptiveness [3].

In conclusion, encoding geometrical information combined with spatial information can improve the descriptiveness of a descriptor, while some existing descriptors only encoding geometrical or spatial information, which have limited descriptiveness. In addition, for encoding geometrical information, the deviation angle between normals or between normal and LRA is commonly used. However, the normal generated on a very small local region [30] has a weak robustness to resist various nuisances. In terms of encoding spatial information, some descriptors (e.g. RoPS, TriSI) do not fully encode the spatial information on LRF by transforming 3D to 2D/1D, and some others are not compact to encode spatial information by repetitive using the information of x, y and z coordinates of local points (e.g. TriSI, TOLDI). Therefore, the descriptiveness and robustness of these existing descriptors have potential for further being improved.

## 3. SDASS-based local shape description

In this section, some techniques (including the construction of Local Reference Axis (LRA), Local Minimum Axis (LMA) and the feature representation of SDASS) involved in our SDASS descriptor are detailed. Specifically, our improved LRA and robust LMA are first detailed. Then, we present the SDASS feature representation by encoding the combination of spatial and geometrical information of a local surface based on the proposed LRA and LMA. Finally, the parameters of our SDASS feature descriptor are quantitatively selected.

### *3.1 Constructing LRA and LMA*

In this section, we construct an LRA developed from the z-axis of the LRF proposed by Yang et al. [23]. To improve the robustness and descriptiveness of encoding geometrical information, we define a new geometrical attribute named Local Minimum Axis (LMA) to replace normal for calculating the deviation angles.

#### *3.1.1 Local reference axis*

In this paper, we select LRA rather than LRF as a basis for spatial division. Although, LRF provides the entire local 3D spatial information including radial, azimuth and elevation directions while LRA loses one-dimension spatial information (i.e., the azimuth information), as shown in Fig.1. However, the repeatability and robustness of the x or y axis are significantly lower than the z axis in an LRF [12, 19], especially in the presence of noise, varying mesh resolutions, etc. To provide a more accurate spatial division in our feature descriptor, we select LRA as a frame to construct our descriptor.

Our LRA is developed from the z-axis in the LRF proposed by Yang et.al. [23]. In below, we first simply repeat the z-axis proposed by Yang et.al. [23], and then propose our improved LRA based on the z-axis.

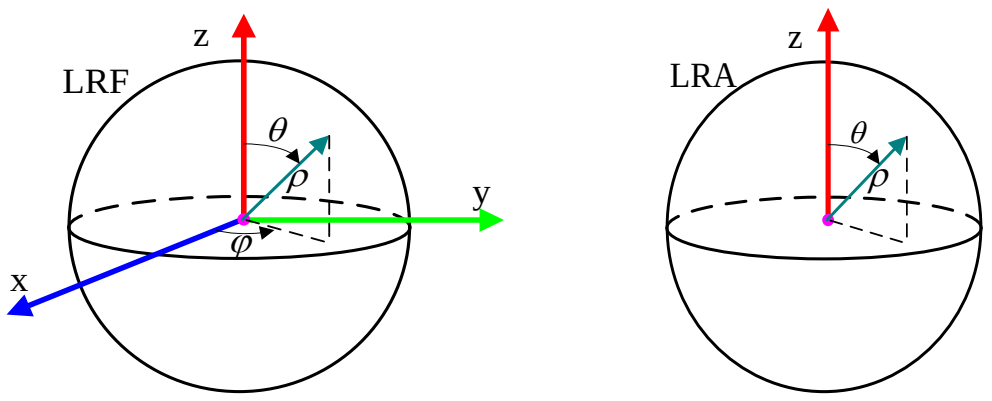

**Fig.1.** LRF and LRA. ($\rho$, $\theta$, $\varphi$) denote the radial, elevation and azimuth directions, respectively [12].

Given a key point $p$ and a support radius R, local points $\mathbf{Q} = \{q_1, q_2, \dots, q_n\}$ can be obtained within the distance of R from the key point $p$. The subset of $\mathbf{Q}$, which is denoted as $\mathbf{Q}_z = \{q_1^z, q_2^z, \dots, q_n^z\}$, determined by 1/3 support radius around the key point is used for calculating the direction of the z-axis. Specifically, based on $\mathbf{Q}_z$, a covariance matrix $Cov(\mathbf{Q}_z)$ is constructed as:

$$Cov(\mathbf{Q}_z) = \begin{bmatrix} q_1^z - \bar{q}^z \\ \dots \\ q_n^z - \bar{q}^z \end{bmatrix}^T \begin{bmatrix} q_1^z - \bar{q}^z \\ \dots \\ q_n^z - \bar{q}^z \end{bmatrix} \tag{1}$$

where $n$ is the size of $\mathbf{Q}_z$, and $\bar{q}^z$ is the centroid of $\mathbf{Q}_z$. The eigenvector $\mathbf{v}(p)$ connected with the minimum eigenvalue of $Cov(\mathbf{Q}_z)$ is calculated as the z axis of this LRF. Next, all support radius neighbors are used for disambiguating the sign of $\mathbf{v}(p)$ as follow:

$$\mathbf{v}(p) = \begin{cases} \mathbf{v}(p), & \text{if } \mathbf{v}(p) \cdot \sum_{i=1}^{n} \mathbf{pq}_i \geq 0 \\ -\mathbf{v}(p), & \text{otherwise} \end{cases} \tag{2}$$

where $\cdot$ between vectors represents dot-product, and $\mathbf{pq}_i$ denotes the vector from $p$ to $q_i$.

As opposed to use the subset of radius neighbors for calculating the direction and using all radius neighbors for disambiguating the sign of the z-axis (also named LRA) proposed by Yang et.al. [23], we use all radius neighbors to determine both the direction and sign of our LRA based on Eqs. (1) and (2). In order to verify the performance of the two strategies (Yang et.al. [23] and our) as well as provide a basis for generating our LMA in Section 3.1.2, a test to the two strategies with respect to different support radii is implemented. For increasing readability, the strategy (Yang et.al. [23]) of using different radii neighbors to respectively calculate the direction and sign of LRA is defined as S1, and the strategy (our) of using same radii neighbors to calculate them is defined as S2. In the strategy S1, a support radius for calculating the direction of LRA is fixed to 20 mesh resolution (hereinafter mr which is computed as the average distance between neighbor points in this paper.) with referring to [12, 31], and a support radius used for eliminating sign ambiguity increases from 3 mr to 20 mr with a step of 1 mr. In the strategy S2, the support radii of calculating the direction and eliminating sign ambiguity are simultaneously increased from 3 mr to 20 mr with a step of 1 mr. In terms of datasets for this test, the three scenes in B3R dataset (the Gaussian noises with the deviation of 0.1, 0.3, 0.5 mr correspondingly combining with the decimation rates of 1/2, 1/4, 1/8) are selected to test the robustness for resisting noise and varying mesh resolutions, and U3OR and U3OR-IR datasets are selected to test the robustness for resisting mesh boundary, clutter and occlusion. The details of these datasets are introduced in Section 4.1.1. The evaluation criterion is introduced in Section 4.1.2. The percentage of angle errors below 5° is counted for evaluating the repeatability of LRA. The results of this test are presented in Fig.2. Several observations can be summarized as follows.

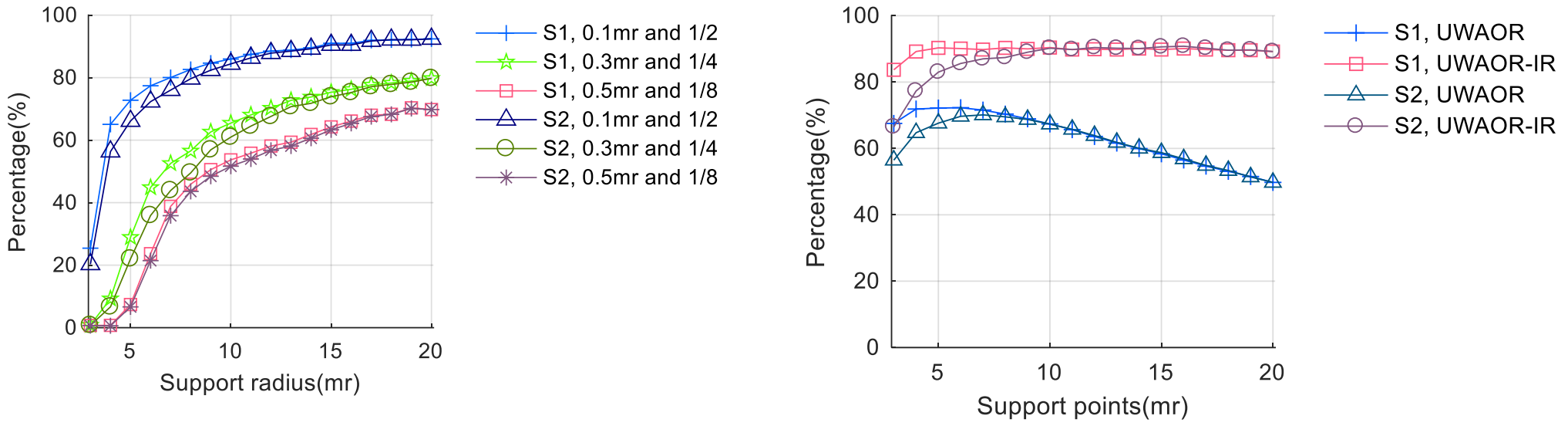

(a) Tested on B3R dataset with the combination of Gaussian noise and varying mesh resolutions.

(b) Tested on U3OR and U3OR-IR datasets.

**Fig. 2.** The repeatability of two strategies (S1 and S2) for generating LRA with respect to different support radii.

*First*, in the presence of noise and varying mesh resolutions (as shown in Fig.2 (a)), the repeatability of the LRA generated by S1 and S2 is gradually improved along with the increase of support radius. Although the repeatability of the LRA generated by S1 is higher than S2 in the process of the support radius increased from 3 mr to 20 mr, the highest repeatability values of them are the same and are appeared with the support radius at 20 mr. So, we can conclude that, for the impact of noise and varying mesh resolutions, the support radii for calculating the direction and eliminating the sign ambiguity of simultaneously taking the maximum value (i.e. 20 mr) can get a high repeatability.

*Second*, for the impact of mesh boundary as tested on U3OR dataset (shown in Fig.2. (b)), the repeatability of the LRA generated by S1 is higher than S2 in the low levels of support radii, and this difference gradually disappear along with the increase of support radius. In addition, the varying tendencies of S1 and S2 tested on U3OR dataset are similar. Their repeatability is gradually improved along with the increase of the support radius before about 6 mr, and then drop after 6 mr. When eliminate the influence of mesh boundary, as tested on U3OR-IR dataset (shown in Fig.2 (b)), the repeatability of the LRA constructed by S1 and S2 are not fall along with the increase of support radius. From this observation, we can conclude that the strategy S1 has a superior performance for resisting the impact of mesh boundary compared to S2, while, when the mesh boundary is eliminated (as tested on U3OR-IR), the best performance of S1 and S2 are similar in the process of support radius increased from 3 mr to 20 mr.

From the above two observations and corresponding conclusions, we can find that our LRA has a strong robustness to noise and varying mesh resolutions, while has a weak robustness to mesh boundary compared to the z-axis proposed by Yang et.al. [23]. There are two reasons of our LRA to improve the robustness to noise and mesh resolution variation while sacrifice the robustness to mesh boundary. The first is that an LRA is difficult to have a comprehensively good performance for resisting various nuisances (e.g. noise, varying mesh resolution and mesh boundary, et. al.). The robustness to noise and mesh resolution variation and the robustness to mesh boundary seem to be contradictory. On the other words, with the increase of support radius in a reasonable range, the robustness to noise and mesh resolution variation is improved while the robustness to mesh boundary is reduced. The second is that mesh boundary is easier eliminated than noise and varying mesh resolutions from a point cloud [11, 25]. So, our proposed LRA is appropriate of being applied on a point cloud with no boundary existent (e.g. the B3R dataset) or boundary identified (e.g. the U3OR-IR dataset). In the case of mesh boundary being existent and not identified, the z-axis of the LRFs (e.g. the LRF proposed by Yang et. al. [23]) with strong robustness to mesh boundary can be used as an LRA to generate our descriptor.

### *3.1.2 Local Minimum Axis*

The deviation angle between normals or between normal and LRA is commonly used for encoding geometrical information in previous descriptors (e.g. FPFH [21], SHOT [24]). In this process, the repeatability of normals determines the performance of encoding geometrical information on a local surface. At present, there are two popular techniques of generating normal. The first technique uses all triangular patches attached with a point to calculate the normal of that point [11, 25, 32]. The second technique first generating a covariance matrix by using the radius neighbors of a point, and then the eigenvector corresponded with the minimum eigenvalue is calculated as the normal of that point [30, 33-34]. In the following, for compactness, the normal generated by the first method is called triangular patches based normal (TN) and the second method is called radius neighbors based normal (RN). However, the disambiguation of the normals constructed by these two techniques commonly rely on a viewpoint [13, 34]. If the viewpoint is unknown, manual to disambiguate of the normals is needed [34], which is inefficiency. In addition, normal is a geometrical attribute to represent a very small local region (e.g. a local surface for generating TN has a support radius of 1 mr averagely, and a local surface for constructing RN usually has a support radius less than 3 mr [30, 35]). The repeatability of the normal generated on a small local region is sensitive to noise, varying mesh resolutions, etc. (as verified in Section 3.1.1).

For overcoming the above two weaknesses existing in normal, we propose a local geometrical attribute called Local Minimum Axis (LMA) to replace the normal for constructing the deviation angle. The method of generating our LMA is similar to the technique of generating the improved LRA introduced in Section 3.1.1. In opposite to use a small region for calculating the direction of normal, the proposed LMA use a larger local region for determining its direction, which significantly increase the robustness to various nuisances. It is worth noting that, since the LMA is generated on a large region, it no longer represents the attribute of normal for a surface, as shown in Fig. 3. According to the tested results presented in Fig. 2, considering the tradeoff among efficiency, the robustness to noise and varying mesh resolutions and the robustness to mesh boundary, the support radius for generating LMA is selected as 7mr in this paper. Based on this support radius, the direction and sign of LMA can be determined by using Eqs. (1) and (2), respectively, which are presented in Section 3.1.1. In this way, disambiguating the sign no longer rely on the view point, which improve the efficiency and accuracy.

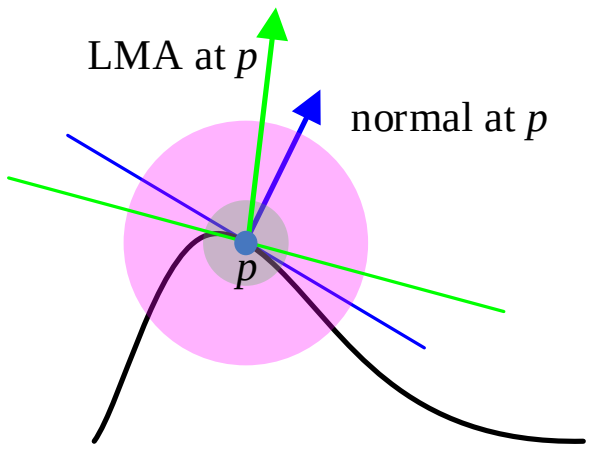

**Fig. 3.** An illustration of the support regions for constructing LMA and normal at point $p$. The larger region represents the support region for generating LMA and the small region denotes the support region for constructing normal.

### *3.2 SDASS descriptor*

Once the LRA and LMA are constructed, the next task is to encode spatial and geometrical information on a local surface. Given a point cloud or surface, the local points are determined by a key point $p$ and a support radius R. We define the local points as $\mathbf{Q}=\{q_1, q_2, \ldots, q_n\}$. As shown in Fig. 4 (b), (c), $\mathbf{Q}$ is first transformed to make $p$ coincided with the coordinate origin, and the LRA of the key point aligned with z axis to achieve the rotation invariance of the local surface. The transformed local points are denoted as $\mathbf{Q}' = \{q_1', q_2', \ldots, q_n'\}$. Then, an appropriate manner of encoding spatial and geometrical information on the local surface is explored.

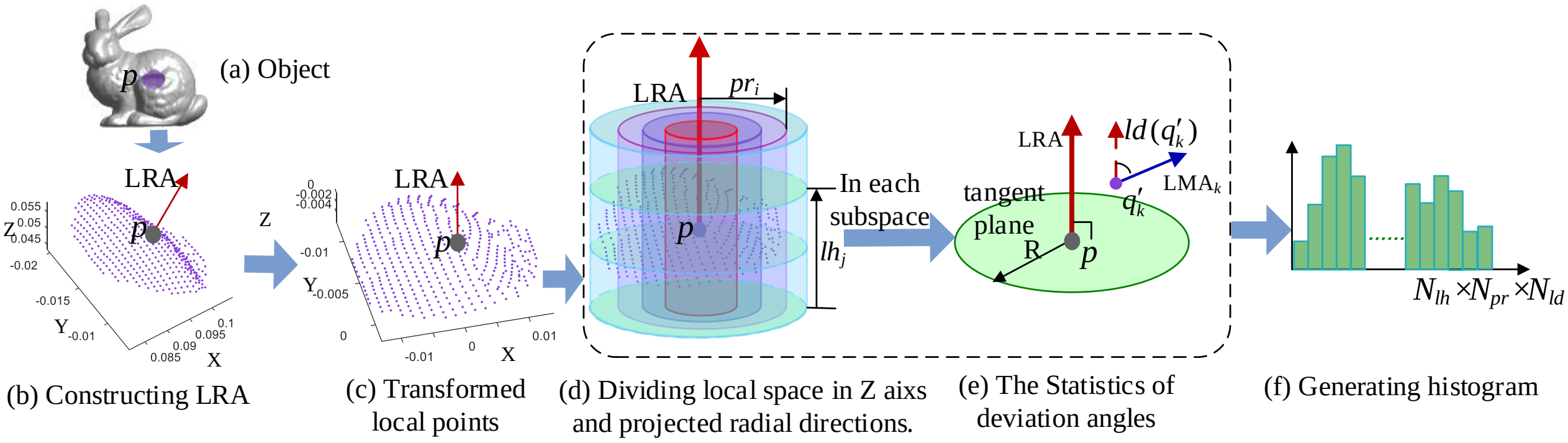

**Fig.4.** An illustration of generating SDASS feature descriptor.

In general, the spatial information of local surface on LRA can be fully encoded by two ways, as shown in Fig.5. One way uses the radial distance ($\rho$) and azimuth angle ($\theta$) to encode spatial information, and the other way uses the height distance ($h$) and projected radial distance ($d$) to encode spatial information. There are some existing descriptors of using the radial distance and azimuth angle to encode spatial information (e.g. SHOT [24], USC [15]), while the majority of existing descriptors are generated by using the height and projected radial metrics (e.g. Spin image [13], TriSI [29], LFSH [3]) to encode spatial information. In this paper, we also use the projected radial and height distance for comprehensively encoding the spatial information on LRA. Besides, our SDASS descriptor uses a geometrical feature (the deviation angle between LRA and LMA) to encode the geometrical information of local surface. The deviation angles between LRA and the normals of neighbors are usually used for encoding geometrical information on local surface, such as SHOT [24], LFSH [3]. Considering that normal represents the attribute of small local region, which is sensitive to various nuisances, we propose a geometrical attribute LMA (see Section 3.1.2), which has obviously higher robustness than normals as verified in Section 4.2.1, to replace the normal for constructing the deviation angles. Next, we detail the feature representation of our SDASS descriptor.

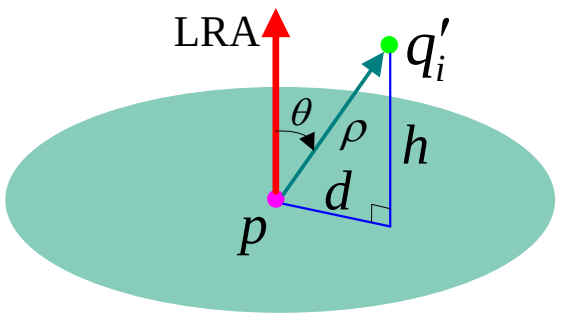

**Fig. 5.** Two ways of encoding spatial information on a local space spanned by LRA

We first encode the spatial information of the transformed local space. The spatial information is encoded by dividing the local space along LRA axis and projected radial direction, as shown in Fig.4 (d). The ranges of divisions in LRA axis and projected radial direction are [0, 2R] and [0, R], respectively. Assume that the numbers of divisions along LRA and projected radial direction are $N_{lh}$ and $N_{pr}$, respectively. Then, the indexes of each transformed local point $q_k'$ falling into which subspace are calculated as follow.

$$\begin{cases} I_{lh} = \left\lceil \left( R + q_k'(3) \right) N_{lh} / (2R) \right\rceil \\ I_{pr} = \left\lceil \left( \sqrt{q_k'(1)^2 + q_k'(2)^2} \right) N_{pr} / R \right\rceil \end{cases} \tag{3}$$

where $I_{lh}$ and $I_{pr}$ denote the indexes of the subspace containing $q_k'$ in LRA and projected radial direction, respectively, and $q_k'(1)$, $q_k'(2)$ and $q_k'(3)$ denote the $x, y, z$ coordinates of $q_k'$, respectively. After performing the divisions of the local space, we next encode the geometrical information of local surface.

In each subspace, the geometrical information is encoded with a statistic of deviation angles between the LRA at key point and the LMA at the neighbors, as shown in Fig.4 (e). The method of constructing LMA has been detailed in Section 3.1.2. For each neighbor point $q_k'$, the deviation angle at this point is calculated as follow.

$$ld(q_k') = \arccos(\mathrm{LRA}, \mathrm{LMA}_k) \tag{4}$$

where $ld(q_k')$ denotes the deviation angle between the LRA at key point and the $\mathrm{LMA}_k$ at neighbor point $q_k'$. The range of deviation angle is [0, $\pi$]. After encoding the geometrical information in a subspace $v_i$, the sub-histogram of statistics to the deviation angles in this subspace is generated, and is called $\mathbf{h}_i$. After generating the sub-histograms of all subspaces, our SDASS descriptor can be constructed by concatenating all the sub-histograms into one histogram denoted as $\mathbf{H} = \{\mathbf{h}_1, \mathbf{h}_2 \ldots, \mathbf{h}_{N_{lh}*N_{pr}}\}$. The length of $\mathbf{H}$ is $N_{lh} \times N_{pr} \times N_{ld}$. To achieve robust to variations of the point density, we normalize the whole descriptor to sum up to 1. In addition, considering that the local space, which is actually a sphere with a radius of R, is regarded as a cylindrical space to be subdivided for encoding spatial information, there may exist some redundant subspaces. To improve the compactness of this descriptor, these redundant bins are eliminated from the histogram $\mathbf{H}$.

In the following, we analyze superiority of the proposed SDASS in theory. Four points are summarized as follows.

*First*, in opposite to use the deviation angle of normals in previous descriptors (i.e. LFSH [3], FPFH [21]), the SDASS use the deviation angle between LMA and LRA for encoding geometrical information of local surface. The LMA has a high repeatability compared to normal (as verified in Section 4.2.1) especially in the presence of various nuisances (e.g. noise, varying mesh resolutions, etc.). The repeatability of LMA/normal determines the accuracy of calculating the deviation angle. Thus, with high repeatability of our LMA, the geometrical information encoded in our SDASS descriptor has strong robustness and high descriptiveness.

*Second*, the SDASS descriptor encode the combination of spatial and geometrical information of a local surface. The spatial information on LRA is completely encoded by dividing the local space along LRA axis and projected radial direction. The geometrical information is encoded by the deviation angles between LRA and LMA. In the view of the components in SDASS descriptor, LFSH is the most similar descriptor to ours. In contrast, the three features used in LFSH are concatenated to constitute the final 1D histogram. Although, it can reduce the dimensions of the descriptor, it is unable to fully encode the spatial information on LRA, and divisional statistics of the geometrical information. In addition, the LFSH directly use the normal at a key point as LRA and encodes local geometrical information by using the deviation angles between the LRA and the normals of neighbors, which greatly reduces the robustness to various nuisances (e.g. noise, varying mesh resolutions, etc.) owing to the low repeatability of normal (as verified in Section 4.2.1).

*Third*, the SDASS descriptor is generated on LRA which usually has high overall repeatability than LRF [12, 19]. We propose an improved LRA used in generating SDASS descriptor. The LRA has a strong robustness to noise and varying mesh resolutions while is sensitive to the boundary of points cloud (as verified in Section 4.2.1). The reason of generating this LRA is for considering that mesh boundary is easier eliminated than noise and varying mesh resolutions from a point cloud [11]. So, we can select to use the improved LRA in the case of a point cloud with no boundary existent or boundary identified. In the case of a point cloud having boundary and not identified, the z-axis of the LRF (e.g. the LRF proposed by Yang et. al. [23]) with high robustness to mesh boundary can be used as the LRA in our descriptor. So, a strong robustness of LRA can guarantee a good performance of the SDASS descriptor.

*Fourth*, the computational efficiency of our SDASS descriptor is high. The computational process of the SDASS descriptor mainly include three steps: transforming local points, dividing local space and implementing the statistics of deviation angles. Obviously, these steps are all simple arithmetic. In addition, unlike some descriptors (e.g. RoPS [22], TriSI [16]) of needing the mesh information of points cloud, the SDASS descriptor is directly applied on original points cloud.

### *3.3 Parameter settings for SDASS descriptor*

There are four parameters to generate our SDASS descriptor. They are respectively the support radius R, the number of subdivisions along LRA axis and projected radial directions (i.e., $N_{lh}$ and $N_{pr}$) and the number of bins for the statistics of deviation angle (i.e., $N_{ld}$). The support radius R is an important parameter because large values of R would affect the computational efficiency and increase the sensitivity to clutter, occlusion and mesh boundary, whereas small values of R would give rise to a descriptor with less distinctiveness [22]. According to the suggestion in [12, 31], 20 mr is selected as the value of support radius R. In this paper, the dimension of SDASS descriptor is $N_{lh} \times N_{pr} \times N_{ld}$-the number of redundant bins. Obviously, the big values of $N_{lh}$, $N_{pr}$, $N_{ld}$ will affect the computational efficiency and consume more memory. If the values are too small, many details on local shape would not be contained in the descriptor. In the following, the process of setting the values for $N_{lh}$, $N_{pr}$, $N_{ld}$ are detailed.

**Table 1** The process of parameter settings for SDASS.

| | $N_{lh}$ | $N_{pr}$ | $N_{ld}$ | $R$ ($mr$) |
|---|---|---|---|---|
| Setting to $N_{lh}$ | 2-20 | 5 | 15 | 20 |
| Setting to $N_{pr}$ | 5 | 2-20 | 15 | 20 |
| Setting to $N_{ld}$ | 5 | 5 | 2-20 | 20 |

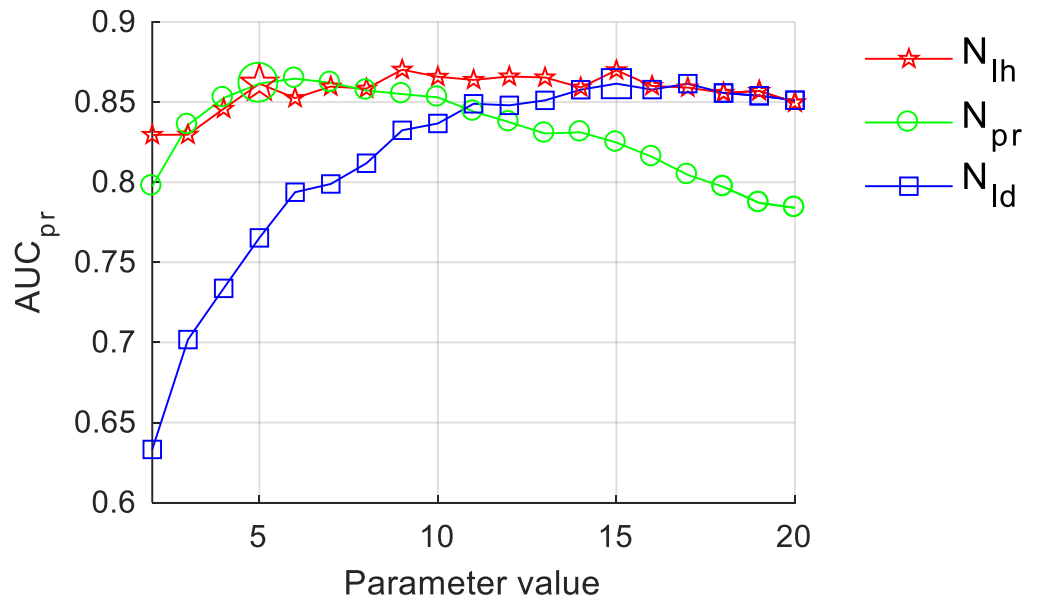

**Fig. 6.** The parameter settings of SDASS descriptor. The larger markers present the selected parameter values.

In this section, we use the scene of combining 1/4 mesh decimation and 0.3 *mr* Gaussian noise in B3R dataset (detailed in Section 4.1.1) to set the parameters of SDASS descriptor. The recall versus 1-precision curve (RPC) [12, 22-23] is a popular method for evaluating the performance of a descriptor. A detailed process of generating RPC curve is presented in Section 4.1.2. For improving compactness, we use the area under the precision-recall curve (called $AUC_{pr}$) to quantitatively evaluate the performance of the SDASS descriptor. In the process of testing the SDASS, the selection of current parameter is on the basis of fixing the other parameters, and the value of the current parameter is set to vary from 2 to 20 with the step of 1. The initial values of $N_{pr}$ and $N_{ld}$ are set to 5 and 15, respectively, with referring to [3]. After current parameter is tested, the value of current parameter can be set, and then it can be used to set the remaining parameters. The detailed process of parameter settings for SDASS is listed in Table 1. The $AUC_{pr}$ results of SDASS with varying $N_{lh}$, $N_{pr}$ and $N_{ld}$ are presented in Fig.6. From the results, we can observe that the $AUC_{pr}$ performance significantly improves as $N_{lh}$, $N_{pr}$ and $N_{ld}$ respectively increased from 2 to 5, from 2 to 5 and from 2 to 15. According to this results combining with the compactness and efficiency of a descriptor, the values of $N_{lh}$, $N_{pr}$ and $N_{ld}$ are set to 5, 5 and 15, respectively.

## 4. Experiments

In this section, the propsoed LRA, LMA and SDASS descriptor are tested on four standard datasets and two synthetic ones for verifying the robustness to noise, varying mesh resolutions, mesh boundary, etc. The four standard datasets include the Bologna 3D retrieval (B3R) dataset [36-37], the UWA 3D object recognition (U3OR) dataset [6], the UWA 3D modeling (U3M) dataset [6, 38] and the Queen's LIDAR (QuLD) dataset [39]. For giving an contrastive evaluation, our SDASS descriptor is compared with some state-of-the-arts (including Spin Image [13], SHOT [24], RoPS [22], TriSI [16], LFSH [3], TOLDI [23]). All tested descriptors are implemented in MATLAB. The MATLAB code of SHOT is kindly provided by author and the MATLAB code of Spin Image, RoPS and TriSI are available in website[2]. The MATLAB codes of LFSH is written by us from its C++ versions which is provided by author. The MATLAB code of TOLDI is written by us with referring to the author's paper [23]. All experiments are conducted on a computer with an Intel Core i7-4790 CPU and 12GB of RAM.

### *4.1 Experimental setup*

In the following, the implementation details of the experiments are introduced. Specifically, we first introduce four standard datasets and two synthetic ones applied in our experiments, and then detail the evaluation criteria for our LRA, LMA and SDASS feature descriptor.

#### *4.1.1 Datasets*

Considering the B3R, U3OR, U3M and QuLD datasets can cover various nuisances (e.g. noise, varying mesh resolutions, clutter and occlusion, etc.), we use these four datasets in this paper for comprehensively evaluating the performance of our descriptor. Some exemplars of the four datasets are shown in Fig.7. Specifically, the B3R dataset contains noise, the U3OR dataset contains occlusions, clutter and mesh boundary, the U3M dataset contains occlusions and mesh boundary, and the QuLD dataset is a quite challenging 3D object recognition dataset, which contains the combination of mesh resolution variation, noise, occlusion, clutter, and mesh boundary. In addition, for comprehensively verifying the robustness to mesh resolution variation and noise, we generate some new scenes in the B3R dataset. For presenting the influence of mesh boundary, we generate two new datasets developed from U3OR and U3M datasets, respectively. The details are presented as follows.

The B3R dataset has 6 models (i.e., "Bunny", "Armadillo", "Asia Dragon", "Happy Buddha", "Dragon" and "Thai Statue") and 18 synthetic scenes. These models are copied from the Stanford 3D Scanning Repository [36], which are obtained by a Cyberware 3030 MS scanner. The 18 scenes in the B3R datasets are created by adding Gaussian noises with the standard deviation of 0.1, 0.3, and 0.5 mr, respectively, in each randomly transformed model. To give a comprehensive comparison for the robustness of descriptors to noise and varying mesh resolutions, three group scenes are generated from the B3R dataset. Each group includes 60 scenes. The first group is generated by adding Gaussian noises with increasing standard deviations from 0.1 to 1.0 mr with an interval of 0.1 mr in each randomly transformed B3R model, which include the 18 scenes copied from original B3R dataset. The second group is created by resampling each randomly transformed model from 10/10 (10/10 is the original resolution) to 1/10 of their original mesh resolution with an interval of 1/10. The third group is generated by correspondingly combining the all levels of Gaussian noise in the first group with the all levels varying mesh resolutions in the second group (e.g. the standard deviation of 0.1 mr combined with the decimation rate of 10/10). The scenes in this dataset without any occlusion or clutter. The purpose of employing this dataset is to verify the robustness to noise and mesh resolution variation.

[2] The website of Spin Image is: http://staffhome.ecm.uwa.edu.au/~00053650/code.html, and the website of RoPS and TriSI is: http://yulanguo.me/.

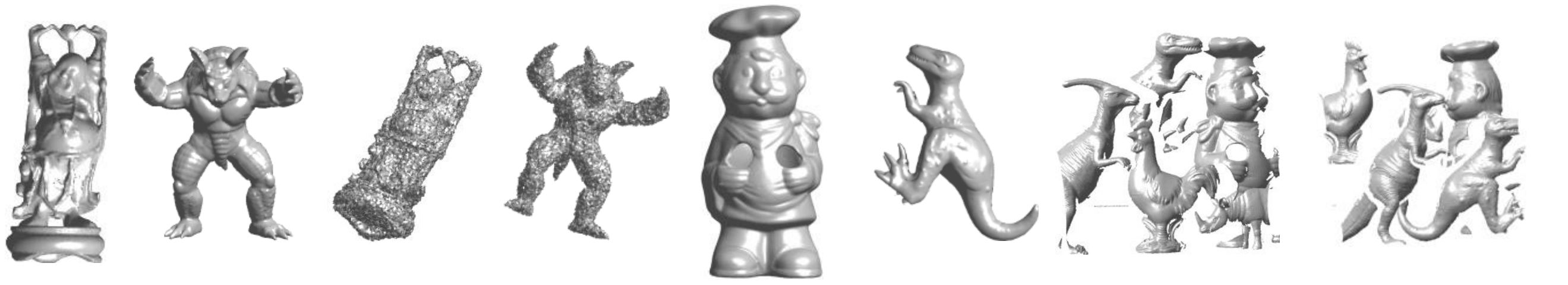

(a) B3R dataset (b) U3OR dataset

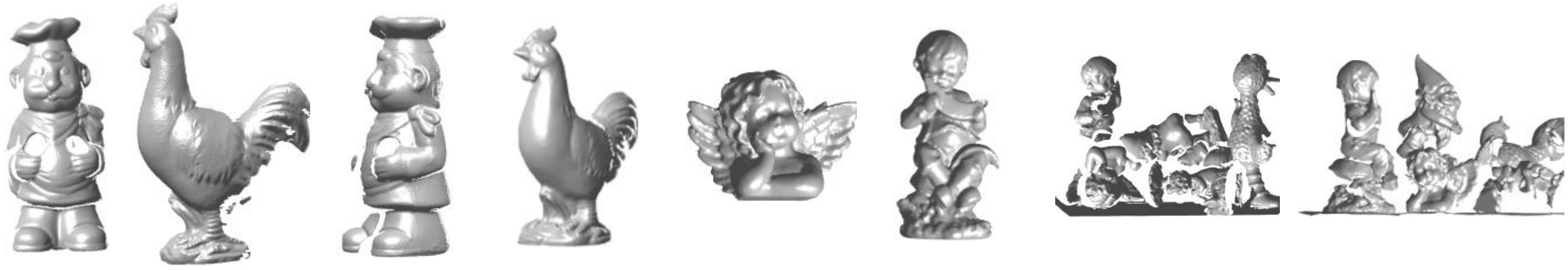

(c) U3M dataset (d) QuLD dataset

**Fig. 7.** Two exemplar models and Two corresponding exemplar scenes respectively taken from the B3R, U3OR, U3M and QuLD datasets.

The U3OR dataset contains 5 models (i.e., “Chicken”, “T -Rex”, “Parasaurolophus”, “Rhino” and “Chef”) and 50 scenes. These scenes are constructed by a Minolta Vivid 910 scanner of scanning some randomly placed models together from a specific viewpoint. In this dataset, clutter, mesh boundary and occlusion are three major challenges. The U3M dataset contains 16, 21, 16, and 22 views which are scanned from four objects (i.e., “Chicken”, “T-Rex”, “Parasaurolophus” and “Chef”) by using a Minolta Vivid 910 scanner, respectively. Since a single viewpoint cannot obtain an entire shape of 3D model, the challenges in this dataset are mesh boundary and self-occlusion, etc. Owing to that every two views of an object in the U3M dataset do not always share an overlap, to guarantee model-scene views of an object having an overlap, five pairs of views from each object with bigger overlap area are selected to test our descriptor (20 view pairs in total).

The QuLD dataset is composed of 5 models (i.e., “Angel”, “BigBird”, “Gnome”, “Kid”and “Zoe”) and 80 scenes. Each scene is generated by placing several models together in a scene and scanned from a specific viewpoint by using a Konica-Minota Vivid 3D scanner [39]. The QuLD dataset is a very challenging dataset [22] due to the following three difficulties. First, the scenes are affected by the noise injected with measurement system. Second, the objects in each scene are highly cluttered. Third, the points in the models are quite sparse and unevenly sampled.

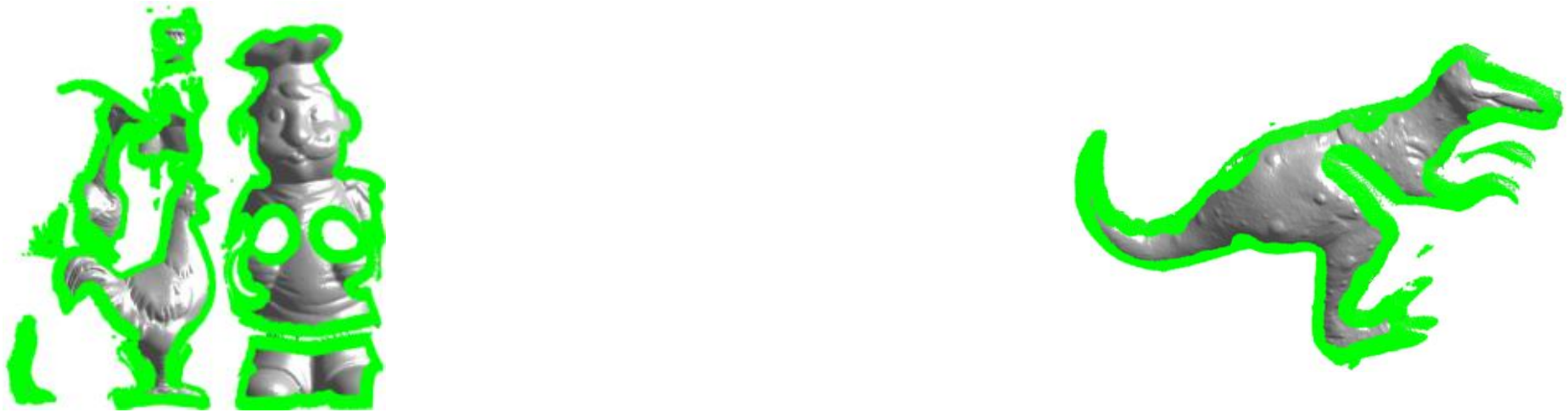

(a) A scene in U3OR-IR dataset (b) A scene or model in U3M-IR dataset

**Fig.8.** An illustration of two scenes for distinguishing boundary and inner regions respectively taken from the U3OR-IR and U3M-IR datasets. The boundary points are shown in green color.

In addition, to present the influence of mesh boundary, we generate two new datasets derived from U3OR and U3M, respectively. For increasing readability, the two modified datasets are named U3OR-IR and U3M-IR, respectively. The new datasets have the same models and scenes with their original datasets. The only difference between them is that the inner and boundary region of the scenes in U3OR-IR and the scenes and models in U3M-IR need to be distinguished. The inner region in a point cloud is defined as the region in which points have a distance larger than the support radius R from the mesh boundary. One exemplar of scene in U3OR-IR and scene or model in U3M-IR are shown in Fig.8. The purpose of distinguishing inner and boundary region is to avoid the influence of mesh boundary by sampling key points only on the inner region, which will be detailed in Section 4.1.2.

### *4.1.2 Evaluation criteria*

We use the recall vs 1-precision curve (RPC) to evaluate the performance of feature descriptors, and directly use angle error to present the repeatability of the axes in LRF, LRA, LMA and normals (TN, RN).

To verify the repeatability of the proposed LRA and LMA, the angle error between two axes is used as an evaluation criterion. The angle error of two arbitrary 3D axes $\mathbf{v}_1$ and $\mathbf{v}_2$ can be simply computed as follow.

$$e = \arccos\left(\frac{\mathbf{v}_1^T \mathbf{v}_2}{\|\mathbf{v}_1\|\|\mathbf{v}_2\|}\right) \tag{5}$$

The RPC is a popular method for evaluating local feature descriptors. The detailed process of generating RPC can refer to [11, 22-23]. If a feature descriptor has a good performance, the recall and precision in its RPC will be high. In order to compactly and quantitatively present the performance of feature descriptors, the area under the precision-recall curve, defined as $AUC_{pr}$, is also used in this paper. $AUC_{pr}$ is a summative and simple metric to evaluate the performance of an algorithm over the whole precision-recall space [11]. In an ideal case, the $AUC_{pr}$ is equal to 1 since the recall is always 1 for any precision.

In our experiments, for the B3R and U3OR datasets, 1000 key points are randomly extracted from scene, and then their corresponding key points are obtained from the models/model by using the ground truth transformation which are provided by the publishers. In terms of the U3M dataset, each pair of model-scene has an overlap. The overlap is determined based upon the ground truth transformation which is got by first manually matching each model-scene pair and then refining them by using ICP algorithm [40-41]. Then, for each model-scene pair, 1000 key points are randomly extracted from the overlap in scene, and their corresponding key points are obtained from the model with the calculated ground truth transformation. For U3OR-IR dataset, the process of selecting key points is similar to that of U3OR except that the 1000 key points on scenes are randomly selected from their inner region. For U3M-IR dataset, the overlapping inner regions on each model-scene pair is identified. The key points in U3M-IR are selected on the overlapping inner regions by the similar method of selecting key points in U3M.

After the key points are determined, if we only evaluate the repeatability of LRF/A, LMA and normals, we just need to generate LRF/A, LMA and normals on these extracted key points, and use the angle errors to evaluate them. If we evaluate the performance of the local feature descriptors, we need to generate the descriptor features on these key points. The RPC curve and the $AUC_{pr}$ are used for evaluating the performance of feature descriptors.

## *4.2 Performance evaluation of the proposed LRA and LMA*

In this section, the repeatability and efficiency of our proposed LRA and LMA are tested. The proposed LRA and LMA are compared with five recent techniques of constructing LRFs and two popular methods for generating normal. The five techniques of LRFs include Mian et al. [25], Guo et al. [22], Petrelli et al. [1], Tomabari et al. [10] and Yang et al. [23]. The two methods for constructing normals are respectively triangular patches based method (TN) [11, 32] and radius neighbors based method (RN) [30, 33], which have been introduced in Section 3.1.2. For convenient expression, LRF, LRA, LMA and normals are collectively called local axes in the following.

### *4.2.1 Repeatability performance*

Three standard datasets (including B3R, U3OR and U3M) and two modified datasets (including U3OR-IR and U3M-IR), which are detailed in Section 4.1.1, are used to test the repeatability of the local axes. The support radius of constructing these local axes are all set to 20 mr for an unbiased comparison. According to the implementation details introduced in Section 4.1.2, we can obtain the angle errors of each method tested on these experimental datasets. The percentage of the angle errors below $5^{o}$ for each method is counted, as shown in Fig.9. In order to compare the repeatability between the x/y and z axis in a LRF, the angle errors of the x and z axes in a LRF are computed respectively.

In the B3R dataset, the robustness of our proposed LRA and LMA to noise and varying mesh resolutions are verified. For conciseness, we select a few of scenes from the scenes of B3R datasets for this experiment. The results of the local axes tested on the selected scenes of B3R dataset are showed in Fig. 9 (a)-(c). Several observations can be drawn from these results. First, the z-axis of Guo et al. and our LRA achieve significantly strong robustness to all levels noises and varying mesh resolutions. In particular, the z-axis of Guo et al. achieves the best performance for resisting varying mesh resolutions, as shown in Fig.9 (b). It may owe to the z-axis proposed by Guo et al. use all the information of local surface rather than only the mesh vertices. It is worth noting that, although the robustness of Guo's z axis to varying mesh resolutions is strong, its efficiency is very low [23]. Our LRA achieves the best performance for resisting the impact of Gaussian noise and Gaussian noise combined with varying mesh resolution, as shown in Fig.9 (a), (c). Second, for the five tested LRFs, the performance of the z axes surpasses the corresponding x axes by a large margin. It implies that the repeatability of an LRA can significantly outperforms that of an LRF since the z-axis in an LRF can be used as an LRA, which is the major reason for us to use an LRA rather than an LRF for constructing our descriptor. Third, for the comparison between our LMA and the two normals (i.e. TN and RN), the performance of our LMA significantly outperforms TN and RN in all the cases, and the gap appears to be more obvious along with the increase of the level of noise and varying mesh solutions. On this basis, our proposed LMA can be integrated in the descriptors (e.g. LFSH, SHOT, etc.), which contain the geometrical information encoded by using normals, for improving their robustness.

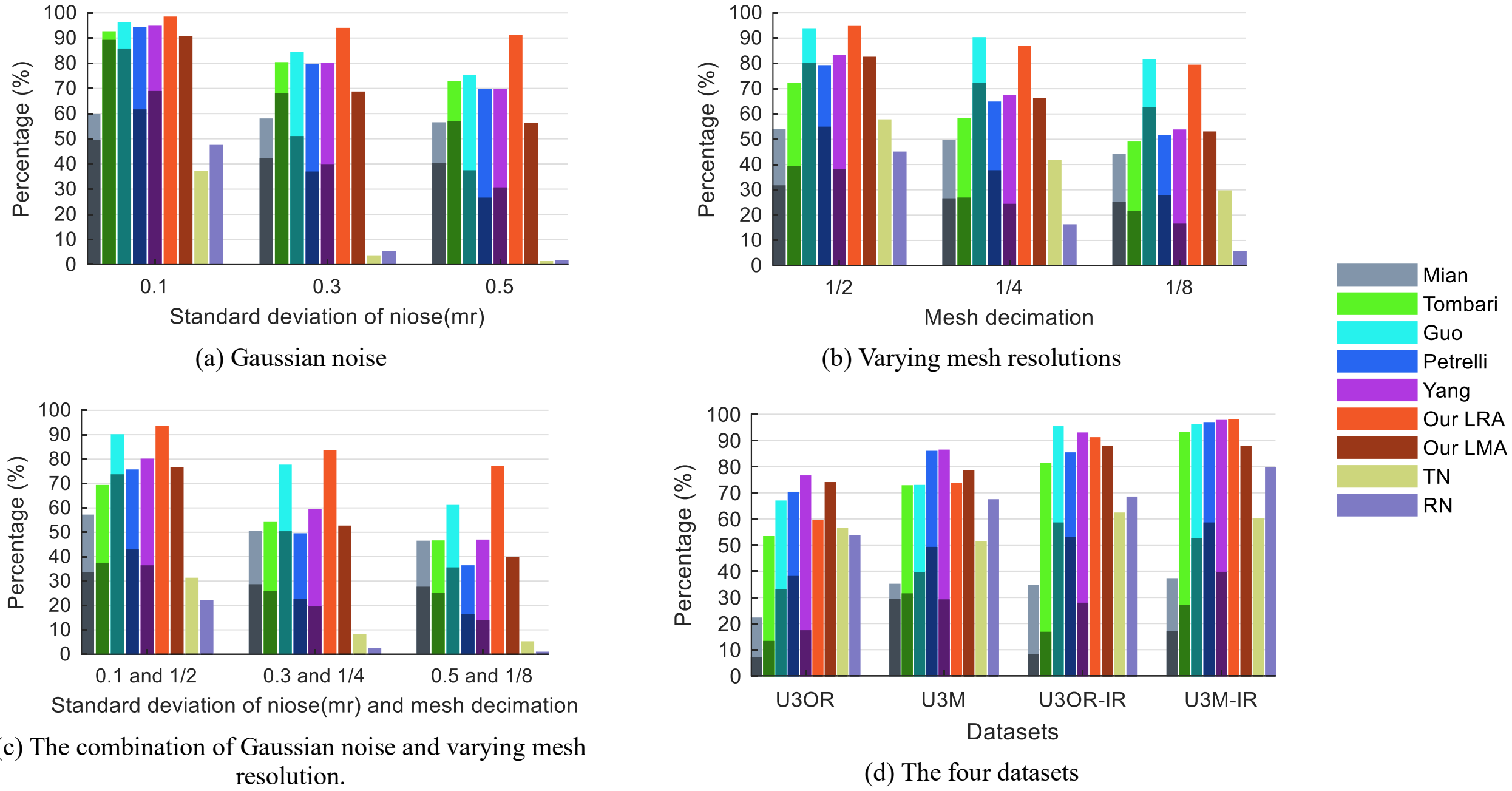

**Fig.9.** The performance evaluation of the repeatability to the local axes (including z and x/y axes in the five LRFs, our LRA, our LMA and two normals). Note that the x and z axis in an LRF are presented in one bin by two colors. The light color denotes the repeatability of its z axis and the corresponding deep color denotes the repeatability of its x axis. Note that the legends of all subfigure are the same with the legend presented in the left of this figure and the number in this legend denote the reference of corresponding method.

In the U3OR and U3M datasets, some different challenges (e.g. mesh boundary, clutter and occlusion) are existent. Besides, in order to present the influence of mesh boundary, the modified datasets U3OR-IR and U3M-IR are used for testing these local axes. The U3OR-IR and U3M-IR are developed from U3OR and U3M, respectively, by identifying the mesh boundary region. The results of testing on the above four datasets are shown in Fig.9 (d). Several observations can be drawn from these results. First, our LMA and the z-axes proposed by Yang et al. [23] and Petrelli et al. [1] are the top three best local axes of testing on U3OR and U3M datasets, while their superior performance are not existent when tested on U3OR-IR and U3M-IR datasets. The thing in common of the above three axes distinguished from the other axes is that the directions of them are all calculated by using a subset of support radius region rather than whole one. So, it is obvious that using a smaller radius to calculate local axes will have strong robustness to mesh boundary. Second, the performance of each method tested on U3OR-IR and U3M-IR correspondingly outperform that tested on U3OR and U3M. It implies that mesh boundary has a great influence to the performance of these local axes. Third, for the comparison of LMA, TN and RN, our LMA outperforms TN and RN by a large margin on all the four datasets. Besides, since TN and RN are generated on a small local region, they have a stronger robustness to resist occlusion, clutter and mesh boundary than noise and varying mesh resolutions. Fourth, Our LRA achieves a good performance on U3OR-IR and U3M-IR datasets while has an inferior performance on U3OR and U3M datasets. It shows that our LRA is vulnerable to the mesh boundary.

In conclusion, our proposed LRA has a strong robustness to noise and varying mesh resolutions, while has a weak robustness to mesh boundary. So, our LRA is appropriate to be applied on a dataset with the boundary being identified or no boundary existent. In the presence of the mesh boundary, the z axis of some state-of-the-arts (such as the z axis proposed by Yang et. al [23]) with robust to mesh boundary can be used as the LRA in our descriptor. Our proposed LMA has a strong robustness to various nuisances (e.g. noise, varying mesh resolution, clutter, occlusion, et al.), and significantly outperforms the two normals (i.e. TN, RN).

### *4.2.2 Time efficiency*

In this section, the efficiency of the methods for generating these local axes (including the six LRFs, our LRA, our LMA, TN and RN) is tested. Since the efficiency is mainly correlate to the number of local points, we only use the B3R dataset to test the efficiency of these local axes. In particular, we first randomly extract 1000 points in each model on the B3R dataset (6000 points in total). For the test of the six LRFs and our LRA, the total time costs of each LRF or our LRA generated on these points with respect to different support radii R are counted. Similar to [23], R also increases from 5 mr to 40 mr with an interval of 5 mr in this paper. For the test of TN, RN and our LMA, since they denote the local geometrical attribute of a surface, we count the total time costs of RN and our LMA on the selected points with a changeless support radius, and the total time cost of PN on the selected points based on the triangular patches connected with these points. The fixed support radius of RN and our LMA are set to 3 mr and 7 mr, respectively.

The time costs of the LRF and our LRA are shown in Fig.10. It can be observed that the computational time of our LRA and the LRF proposed by Mian et al. [25] and Tombari et al. [10] are similar, and they achieve the best performance compared to the others. The LRF proposed by Guo et al. [22] has the worst performance of computational efficiency, and its performance is obviously inferior than the others. It is because that the LRF is generated based on a whole local surface rather than the vertices on a local surface. The LRF proposed by Yang et al. [23] and Petreli et al. [1] achieve a medium performance in terms of time efficiency. For the comparison of TN, RN and our LMA, the computational times of TN, RN and our LMA generated on the 6000 extracted points are 0.0924s, 0.2820s and 0.3618s, respectively. We can see that the computational efficiency of TN outperforms RN and LMA, while the computational efficiency of TN and our LMA are comparable. It is mainly because that RN and our LMA need to calculate a covariance matrix while TN do not need to. Although our LMA has the worst computational efficiency compared with TN and RN, the gaps of the computational time among them are not obvious. Besides, our LMA is directly generated on a point cloud, while constructing the TN need the information of triangular mesh.

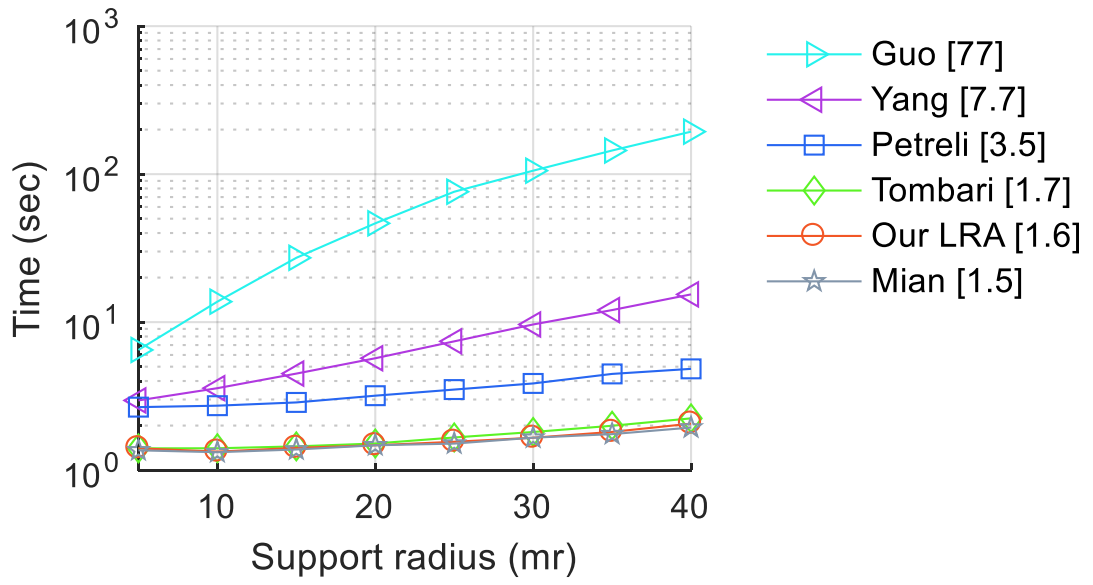

**Fig.10.** The efficiencies of the five LRF and our LRA. The Y axis is shown logarithmically for clarity. The numbers in square brackets are the average computational time and listed in a descending order.

## *4.3 Performance evaluation of the SDASS*

In this section, the proposed SDASS descriptor is evaluated on the four datasets (i.e., B3R, U3OR, U3M and QuLD) and two modified datasets (i.e., U3OR-IR and U3M-IR) by using both the RPC curve and $AUC_{pr}$ (introduced in Section 4.1.2), and compared with six state-of-the-arts for a thorough evaluation. The six descriptors include the SI [14], SHOT [10], RoPS [22], TriSI [29], LFSH [3] and TOLDI [23]. For these six selected descriptors, SI is the most cited one for describing local shape, SHOT, RoPS and TriSI are verified with an advanced performance [11, 23], and LFSH and TOLDI are recently proposed descriptors. The parameter settings of these seven descriptors are presented in Table 2. Note that the parameter settings of our SDASS is composed of $N_{lh} \times N_{pr} \times N_{ld}$ -the number of redundant bins.

To verify the performance of the proposed LMA, we replace the normal used in SI, SHOT and LFSH with our LMA, and then generate three modified descriptors: SI combined with the proposed LMA (called SI+LMA), SHOT combined with the proposed LMA (called SHOT+LMA), and LFSH combined with the proposed LMA (called LFSH+LMA). Note that the parameter settings in SI+LMA, SHOT+LMA and LFSH+LMA are the same with their original descriptors. In addition, for satisfying some time-crucial applications (e.g. robotics, mobile phones, etc.), the comparison of these descriptors in terms of computing efficiency is also presented.

**Table 2** Parameter settings for seven descriptors, where mr denotes mesh resolution.

| | Support radius (mr) | Dimensionality | Length |
|---|---|---|---|
| SI | 20 | 15×15 | 64 |
| SHOT | 20 | 8×2×2×11 | 352 |
| RoPS | 20 | 3×3×3×5 | 135 |
| TriSI | 20 | 15×15×3 | 675 |
| LFSH | 20 | 15+10+5 | 30 |
| TOLDI | 20 | 3×20×20 | 1200 |
| SDASS | 20 | 15×5×5-30 | 345 |

### *4.3.1 Performance on the B3R Dataset*

The purpose of testing on the B3R dataset is mainly to verify the robustness to noise and mesh resolution variation. For each descriptor, we follow the procedures introduced in Section 4.1.2 to construct RPC curve and $AUC_{pr}$. The results of all tested descriptors are presented in Fig.11. In particular, the left of Fig.11 shows the RPC results of one typical case for each nuisance (including Gaussian noise, mesh resolution variation and the combination of Gaussian noise and varying mesh resolutions) and the right of Fig.11 present the $AUC_{pr}$ results of all levels for the corresponding nuisance.

In the view of robustness to noise (Fig.11 (a) and (b)), several observations can be given as follows. First, the performance of our SDASS descriptor surpasses that of the others by a large margin under all levels of noise and the advantage of our descriptor is more obvious with the increase of noise levels. Second, the descriptor SHOT+LMA achieves the second best performance for resisting noise, and SI and LFSH have the worst performance for resisting noise. Third, as the noise levels increased, the performance of SI, LFSH and SHOT are deteriorated sharply. The descriptors SI+LMA, LFSH+LMA and SHOT+LMA correspondingly outperform their original descriptors SI, LFSH and SHOT by a large margin. The main reason for this phenomenon is that the robustness of our LMA to noise significantly outperforms the normals (as verified in Section 4.2.1).

In terms of robustness to varying mesh resolutions, as shown in Fig.11 (c), (d), several observations can be made from the results. First, our SDASS descriptor outperforms all the other descriptors under all levels of mesh decimation, and the gap is broadened with the increase of mesh decimation. Second, in the range of decimation rate from 10/10 to 4/10, the performance of TriSI descriptor is close to our SDASS descriptor. As the levels of decimation rate surpass 4/10, the performance of TriSI deteriorates sharply and the gap between our SDASS and TriSI is more obvious. Similar to TriSI, the performance of RoPS and TOLDI also have a significant drop in the high levels of decimation rate. The common trait of TriSI, RoPS and TOLDI is that they are generated by a view-based mechanism. Third, similar to the performance of resisting noise, the descriptors SI+LMA, LFSH+LMA and SHOT+LMA also outperform their original descriptors SI, LFSH and SHOT in terms of resisting varying mesh resolutions, respectively, which further verify the superiority of LMA.

For the robustness to the impact of combining noise and varying mesh resolutions, as shown in Fig.11 (e), (f), our SDASS descriptor also significantly outperform the other methods in all levels of combined noise and mesh decimation (except the highest level). In the highest level of this nuisance, all descriptors fail to work, as shown in Fig.11 (f). The overall performance of SHOT+LMA is the second best. The TriSI has a better performance in the low levels of combined noise and mesh decimation while the superiority is no longer existent in the high levels. The SHOT+LMA, TOLDI, LFSH+LMA, RoPS,

SHOT and TriSI achieve a moderate performance, and their performance are comparable. The performance of SI and LFSH are significantly inferior to the others. Their fail to work even under the low level of this nuisance. It is because that the SI and LFSH descriptors use the normal, which has a weak robustness to noise and mesh resolution variation, as their LRA.

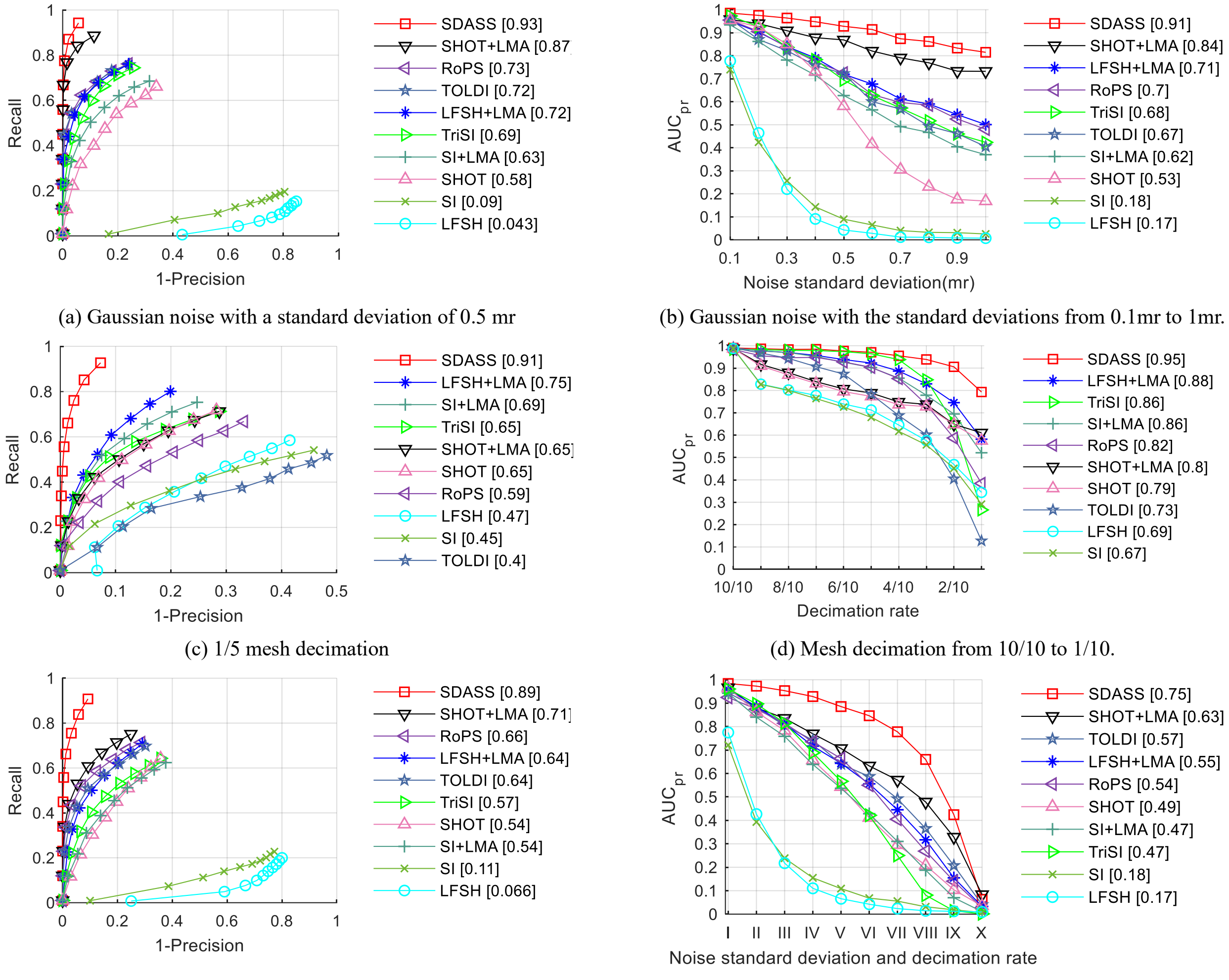

(a) Gaussian noise with a standard deviation of 0.5 mr

(b) Gaussian noise with the standard deviations from 0.1mr to 1mr.

(c) 1/5 mesh decimation

(d) Mesh decimation from 10/10 to 1/10.

(e) 0.5 mr Gaussian noise and 1/2 mesh decimation

(f) The Gaussian noises in subfigure (b) correspondingly combined with the decimation rates in subfigure (d).

**Fig.11.** The RPC and $AUC_{pr}$ evaluation of the ten descriptors tested on the B3R dataset. The *x* labels in the subfigure (f) present the Guassian noises in the subfigure (b) correspondingly combined with the decimation rates in the subfigure (d). In the subfigures (a), (c) and (e), the numbers in square brackets are the $AUC_{pr}$ values computed over the corresponding precision vs recall curves. In the subfigures (b), (d), (f), the numbers in square brackets are the mean $AUC_{pr}$ values computed over the corresponding curves. These numbers are listed in a descending order.

### *4.3.2 Performance on U3OR, U3OR-IR, U3M and U3M-IR Datasets*

In this section, all descriptors are tested on U3OR, U3OR-IR, U3M and U3M-IR datasets for verifying the robustness to clutter, occlusion and mesh boundary. The propose of using U3OR-IR and U3M-IR datasets is mainly to present the influence of mesh boundary to the descriptors. The U3OR-IR and U3M-IR datasets are introduced in Section 4.1.1. In addition, in consideration of the LRF proposed by Yang et.al [23] has the best performance to resist occlusion, clutter and mesh boundary which are existent in U3OR and U3M datasets (see Section 4.2.1), we also use the z-axis in this LRF to replace our LRA for generating the new SDASS descriptor, and call this new integrated descriptor "SDASS-Yang" in the following.

The results of all descriptors tested on U3OR, U3M, U3OR-IR and U3M-IR datasets are shown in Fig.12. Several observations can be drawn from these results. First, the descriptors SDASS-Yang and SDASS achieve the best performance on all four datasets (i.e., U3OR, U3M, U3OR-IR and U3M-IR), and they outperform the others by a large margin. In particular, SDASS-Yang slightly outperform SDASS on U3OR and U3M datasets, and they have a similar performance on U3OR-IR and U3M-IR datasets. It is mainly because that the performance of z-axis in the LRF proposed by Yang et.al [23] outperforms our LRA when tested on U3OR and U3M, and they have a similar performance when tested on U3OR-IR and U3M-IR. Second, the performance of all descriptors is universally improved from tested on U3OR and U3M to U3OR-IR and U3M-IR. It implies that mesh boundary has a significant impact to the performance of the descriptors. Third, similar to the results tested on B3R dataset, the performance of SI+LMA, SHOT+LMA and LFSH+LMA is significantly improved by combining our LMA with their original descriptors SI, SHOT and LFSH. It implies that the proposed LMA has a higher robustness to clutter, occlusion and mesh boundary compared to the normals (i.e., TN, RN).

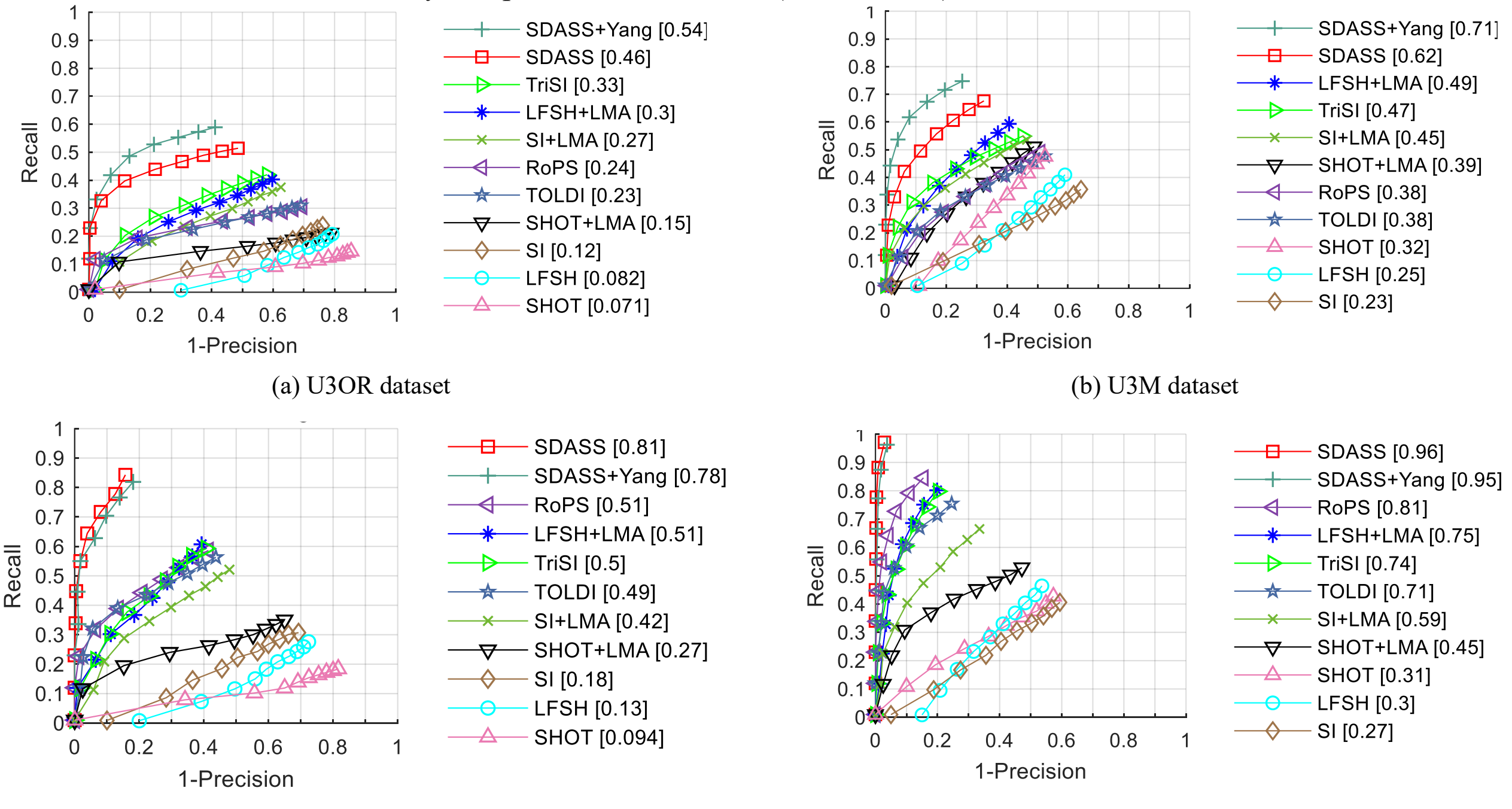

(a) U3OR dataset (b) U3M dataset

(c) U3OR-IR dataset (d) U3M-IR dataset

**Fig.12.** The RPC and $AUC_{pr}$ evaluation of the eleven descriptors tested on the four datasets (U3OR, U3M, U3OR-IR, U3M-IR). The numbers in square brackets are the $AUC_{pr}$ values computed over the corresponding precision vs recall curves, and listed in a descending order.

### *4.3.3 Performance on QuLD Dataset*

In this section, the eleven descriptors (including SDASS+Yang) are tested on QuLD dataset. The QuLD dataset is a very challenging dataset. The nuisance contained in this dataset is the combination of noise, mesh resolution variation, clutter, occlusion and mesh boundary. The results of the eleven descriptors tested on this dataset are shown in Fig. 13. From the results, we can see that the descriptors SDASS+Yang and SDASS achieve the best performance, and their performance obviously outperform the performance of the others. The performance of SDASS+Yang slightly surpass the performance of SDASS, which is because that the LRA proposed by Yang et al. [23] has a higher repeatability than our LRA in case of mesh boundary existed in a dataset. The descriptors SI, LFSH and SHOT have the worst performance. It is mainly because that the normal used in these descriptors present a low repeatability in the impact of various nuisances (including noise, varying mesh resolutions, et al.) which are all contained in the QuLD dataset.

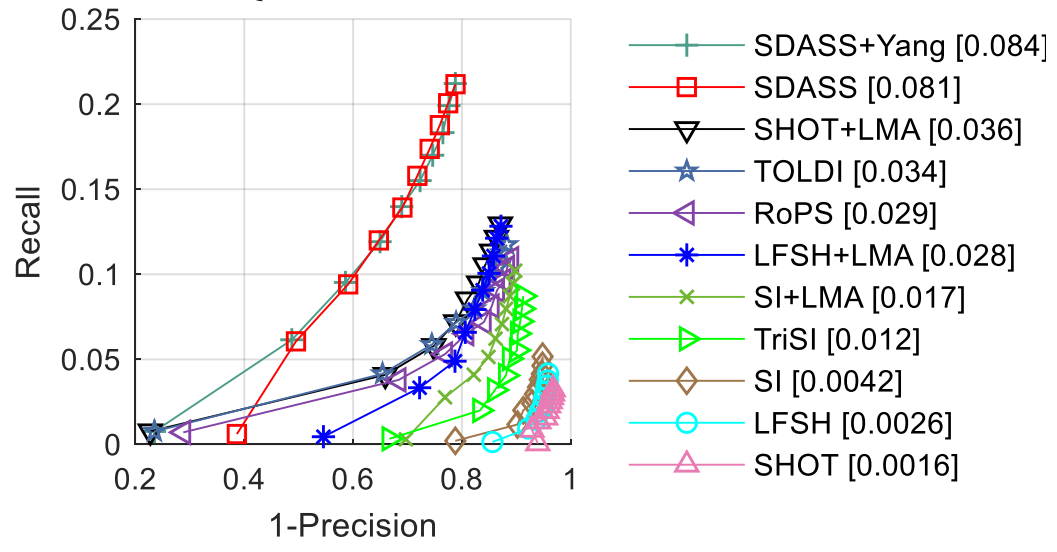

**Fig.13.** The RPC and $AUC_{pr}$ evaluation of the eleven descriptors tested on the QuLD dataset. The numbers in square brackets are the $AUC_{pr}$ values computed over the corresponding precision vs recall curves, and listed in a descending order.

#### 4.3.4 *Time efficiency*

In this section, the efficiency of the seven descriptors (including SI, SHOT, RoPS, TriSI, LFSH, TOLDI and SDASS) are tested. Since the efficiency is mainly correlate to the number of local points, we just use the B3R dataset to test these descriptors. Specifically, 1000 points in each model of B3R dataset are randomly extracted (6000 points in total). Then, the total time costs of each descriptor implemented on the extracted points with different support radii R are counted. Here, the values of R are the same with the set in Section 4.2.2. The tested results of these descriptors are shown in Fig.14.

For the results shown in Fig.14, the efficiencies of LFSH, SI, SDASS, TOLDI and SHOT are comparable. Specifically, among these descriptors, LFSH and SI rank the first compared to all the other descriptors, SDASS and TOLDI achieve the second best of calculation efficiency, and SHOT has the worst performance among these five descriptors. The computational performance of TriSI and RoPS are the worst and inferior than the others about one order magnitude. In conclusion, our SDASS achieve a better computational performance although it is slightly inferior to the computational performance of LFSH and SI. It is worth noting that the efficiencies of RoPS and TriSI descriptors are the lowest, and the times of generating them are mainly spent in the construction of the LRFs (see Section 4.2.2).

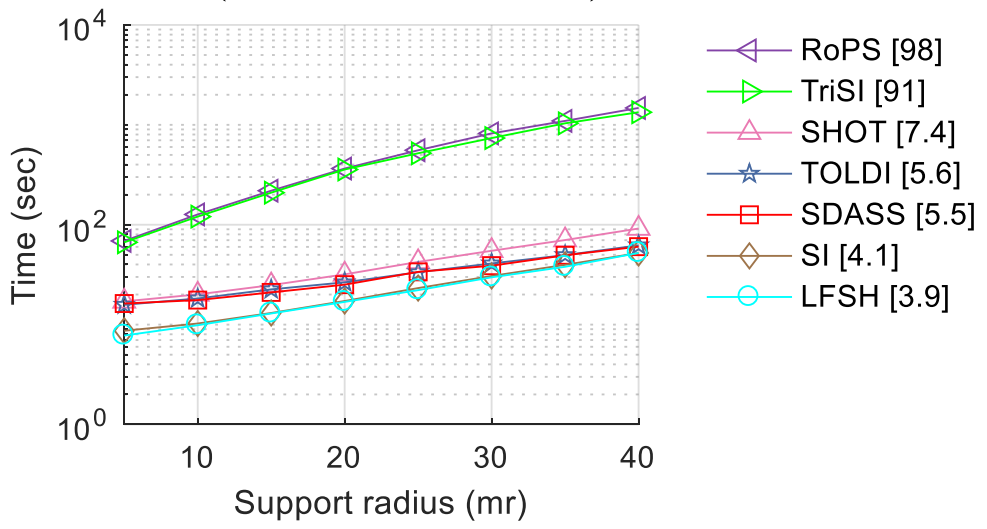

**Fig.14.** The computational efficiencies of the seven descriptors with respect to different support radii. The Y axis is shown logarithmically for clarity. The numbers in square brackets are the average computational time, and are listed in a descending order.

## 5. Applications to 3D matching

In this section, we evaluate the performance of our SDASS descriptor in the application of 3D matching, and compared with the ten feature descriptors (including SI, SHOT, RoPS, TriSI, LFSH, TOLDI, SI+LMA, SHOT+LMA, LFSH+LMA and SDASS+Yang). The performance of these descriptors in 3D matching are evaluated by the percentage of correct correspondences (PCC) [31]. Specifically, we first randomly sample some key points on scene and 1000 key points on model. Then, the 11 feature descriptors are generated on these key points. For each model feature vector, the closest scene feature vector is searched to construct feature correspondence by using *k*-d tree method [42]. Finally, the top 200 feature correspondences with minimum $L_2$ norm distance are selected to calculate the PCC value. In this experiment, 16 pairs of scanned point clouds are used for this assessment. From these point clouds, 8 pairs (2 pairs in each of “Bunny”, “Dragon”, “Happy Buddha” and “Armadillo”) are taken from the Stanford Scanning Repository [36] and 8 pairs (2 pairs in each of “Chef”, “Chicken”, “Parasaurolophus” and “T-Rex”) are taken from the U3M dataset [38]. In this application, the ground truth transformation of these point cloud pairs is known a prior. The PCC can be determined with these ground truth transformations.

The PCC results of the 16 point clouds pairs achieved by our SDASS descriptor are presented in Fig.15. In Fig.15, the results of 3D matching are achieved by using the isometry-enforcing game proposed in [43] based on the 200 feature correspondences for each pair of the point clouds. We can see that the 16 pairs of point clouds can be achieved accurate 3D match based on the correspondences generated by our SDASS descriptor. The PCC results of all the 11 descriptors are presented in Table 3. From Table 3, we can observe that the overall performance of SDASS+Yang is the best, and followed by SDASS descriptor. It is mainly because that the LRA proposed by Yang et al. [23] has a higher robustness than our LRA to resist the influence of mesh boundary which is existed in these point clouds. The rank of these results is almost same with the rank of RPC results presented in Section 4.3.2.

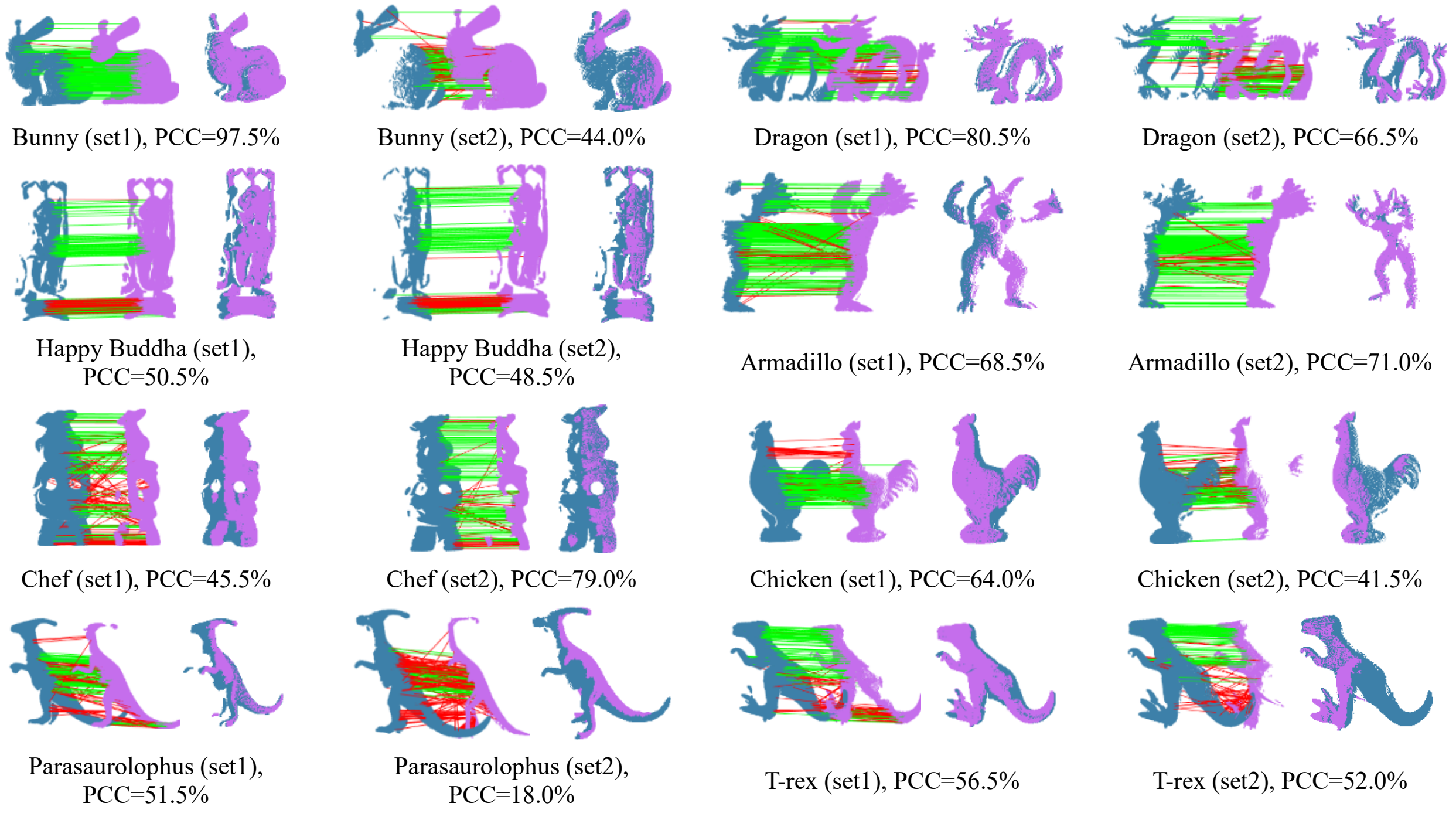

**Fig. 15.** The 16 pairs of point clouds and the PCC results of our SDASS descriptor. The results of 3D matching are achieved by the technique proposed in [43] based on the generated correspondences.

**Table 3** The PCC results of the eleven feature descriptors tested on the 16 pairs of scanned point clouds. The best result is reported in bold and italic face, and the second and third best results for each case are shown in bold face.

| Cases | SI | SI+LMA | SHOT | SHOT +LMA | RoPS | TriSI | LFSH | LFSH+ LMA | TOLDI | SDASS | SDASS +Yang |
|---|---|---|---|---|---|---|---|---|---|---|---|
| Bunny (set1) | 27.0% | 35.5% | 5.0% | 42.5% | **81.5%** | 77.0% | 20.5% | 58.5% | 73.0% | ***97.5%*** | **95.5%** |
| Bunny (set2) | 16.0% | 21.0% | 2.5% | 7.5% | 13.0% | 30.5% | 13.5% | 28.5% | **33.0%** | **44.0%** | ***45.0%*** |
| Dragon (set1) | 49.0% | 61.5% | 43.0% | 56.5% | 45.5% | **77.5%** | 36.0% | 61.5% | 51.5% | **80.5%** | ***84.0%*** |
| Dragon (set2) | 42.5% | 55.0% | 33.5% | 47.5% | 46.0% | **62.5%** | 23.5% | 52.0% | 38.0% | **66.5%** | ***72.5%*** |
| Happy Buddha (set1) | 22.0% | 29.0% | 22.5% | 28.0% | 46.0% | 42.5% | 36.0% | **47.5%** | 41.0% | **50.5%** | ***51.0%*** |
| Happy Buddha (set2) | 27.0% | 31.0% | 23.5% | 26.0% | **44.5%** | 40.5% | 33.5% | 43.0% | 43.0% | ***48.5%*** | **45.5%** |
| Armadillo (set1) | 29.0% | 41.0% | 21.0% | 37.0% | 36.0% | **49.5%** | 20.0% | 46.0% | 27.0% | **68.5%** | ***81.5%*** |
| Armadillo (set2) | 42.0% | 54.0% | 22.5% | 49.0% | 33.5% | **61.0%** | 20.0% | 49.0% | 36.0% | **71.0%** | ***86.0%*** |
| Chef (set1) | 20.5% | 25.0% | 1.0% | 2.0% | 3.0% | 25.0% | 13.5% | **30.0%** | 11.5% | **45.5%** | ***52.5%*** |
| Chef (set2) | 31.0% | 35.0% | 2.5% | 13.5% | 24.5% | 44.5% | 43.5% | **61.5%** | 36.0% | **79.0%** | ***83.5%*** |
| Chicken (set1) | 22.5% | 19.0% | 2.0% | 2.5% | 31.0% | 30.0% | 15.5% | 33.0% | **44.0%** | ***64.0%*** | **57.0%** |
| Chicken (set2) | 26.5% | 28.5% | 0.0% | 1.0% | 8.0% | 28.5% | 19.5% | **31.5%** | 25.0% | **41.5%** | ***52.0%*** |
| Parasaurolophus (set1) | 22.5% | 27.5% | 2.5% | 10.5% | 27.0% | 35.5% | 13.5% | 26.5% | **40.5%** | **51.5%** | ***54.0%*** |
| Parasaurolophus (set2) | 11.5% | **14.0%** | 0.5% | 4.0% | 7.5% | 12.0% | 5.0% | 9.5% | 9.5% | **18.0%** | ***22.5%*** |
| T-rex (set1) | 30.0% | 31.0% | 12.5% | 21.5% | 30.0% | 42.0% | 27.5% | **44.5%** | 36.5% | **56.5%** | ***71.0%*** |
| T-rex (set2) | 21.5% | 24.5% | 26.0% | 31.0% | 33.5% | 31.0% | 32.0% | **42.0%** | 30.5% | **52.0%** | ***58.5%*** |
| Mean PCC | 27.5% | 33.3% | 13.8% | 23.8% | 32.0% | **43.1%** | 23.3% | 41.5% | 36.0% | **58.4%** | ***63.3%*** |

## 6. Conclusions and future work

In this paper, we proposed a novel SDASS feature descriptor for 3D local surface description. The prominent advantage of our SDASS descriptor is its high descriptiveness and strong robustness. Moreover, our SDASS has superior performance in terms of compactness and efficiency. For generating our SDASS descriptor, we also proposed an improved LRA and a robust LMA.

Our LRA is developed from the LRF proposed by Yang et.al. [23]. It achieved a high repeatability in the presence of noise and mesh resolution variation compared to existing techniques. Our proposed LMA achieved superior performance for resisting various nuisances (e.g. noise, varying mesh resolutions) compared to the two methods of constructing normals (TN and RN). Our SDASS descriptor is generated by encoding the combination of spatial and geometrical information of a local surface. The spatial information is fully encoded by subdividing local space along two directions on an LRA. The geometrical information is encoded by using the deviation angle between LMA and LRA. Owing to the strong robustness and high descriptiveness of LMA, the geometrical information of local surface encoded in our SDASS descriptor presented a superior performance. We performed a set of experiments to assess our SDASS descriptor with respect to a set of different nuisances including noise, varying mesh resolutions and mesh boundary, etc. The experimental results showed that our SDASS descriptor outperforms the state-of-the-art methods by a large margin, and obtains high descriptiveness and strong robustness to resist noise, varying mesh resolutions and other variations. At last, our SDASS together with several state-of-the-arts were applied in 3D registration. Our SDASS descriptor achieved the best performance, which further verify the superiority of our SDASS descriptor.

In the future, several interesting research directions will be further explored. Since local feature description is a fundamental task in 3D computer vision, further exploring an effective method of constructing superior descriptor is valuable. In addition, along with the development of some low-cost instruments, e.g., Kinect and stereo sensors, both geometric and photometric cues of objects can be easy obtained. Thus, integrating RGB information in our SDASS descriptor is helpful for applying on the models with limited geometric features while rich photometric cues.

## Acknowledgment

We acknowledge the Stanford University, the University of Western Australia (UWA) and the Queen’s University for providing the datasets. We also acknowledge the help of Dr. Guo, Dr. Yang and Dr. Tombari for providing their codes to us. This work was partly supported by the National Natural Science Foundation of China (Project No. 51575354), and the Shanghai Municipal Science and Technology project (Project No. 16111106102).

## References

[1] A. Petrelli, L. Di Stefano, Pairwise Registration by Local Orientation Cues, Computer Graphics Forum, 35 (2016) 59-72.

[2] Y. Guo, F. Sohel, M. Bennamoun, J. Wan, M. Lu, An Accurate and Robust Range Image Registration Algorithm for 3D Object Modeling, IEEE Transactions on Multimedia, 16 (2014) 1377 - 1390.

[3] J. Yang, Z. Cao, Q. Zhang, A fast and robust local descriptor for 3D point cloud registration, Information Sciences, 346-347 (2016) 163-179.

[4] Y. Guo, B. Mohammed, S. Ferdous, M. Lu, J. Wan, 3D Object Recognition in Cluttered Scenes with Local Surface Features: A Survey, IEEE Transactions on Pattern Analysis and Machine Intelligence, 36 (2014).

[5] A. Aldoma, F. Tombari, L.D. Stefano, M. Vincze, A Global Hypothesis Verification Framework for 3D Object Recognition in Clutter, IEEE Transactions on Pattern Analysis and Machine Intelligence, 38 (2016) 1383-1396.

[6] A.S. Mian, M. Bennamoun, R. Owens, Three-Dimensional Model-Based Object Recognition and Segmentation in Cluttered Scenes, IEEE Transactions on Pattern Analysis and Machine Intelligence, 28 (2006) 1581-1601.

[7] A.M. Bronstein, M.M. Bronstein, L.J. Guibas, M. Ovsjanikov, Shape Google: Geometric Words and Expressions for Invariant Shape Retrieval, ACM Transactions on Graphics, 30 (2011) 1-20.

[8] Y. Guo, Q. Dai, View-Based 3D Object Retrieval: Challenges and Approaches, IEEE Multimedia, 21 (2014) 52-57.

[9] Y. Lei, M. Bennamoun, M. Hayat, Y. Guo, An efficient 3D face recognition approach using local geometrical signatures, Pattern Recognition, 47 (2014) 509-524.

[10] F. Tombari, S. Salti, L. Di Stefano, Unique Signatures of Histograms for Local Surface Description, European Conference on Computer Vision, Springer, New York, 2010, pp. 356-369.

[11] Y. Guo, M. Bennamoun, F. Sohel, M. Lu, J. Wan, N.M. Kwok, A Comprehensive Performance Evaluation of 3D Local Feature Descriptors, International Journal of Computer Vision, 116 (2016) 66-89.

[12] J. Yang, Q. Zhang, Z. Cao, The effect of spatial information characterization on 3D local feature descriptors: A quantitative evaluation, Pattern Recognition, 66 (2017) 375-391.

[13] A.E. Johnson, Spin-Images: A Representation for 3-D Surface Matching, Carnegie Mellon University, 1997.

[14] A.E. Johnson, M. Hebert, Using Spin-Images for Efficient Object Recognition in Cluttered 3-D Scenes, IEEE Transactions on Pattern Analysis and Machine Intelligence, 21 (1999).

[15] F. Tombari, S. Salti, L. Di Stefano, Unique Shape Context for 3D Data Description, ACM Workshop on 3D Object

Retrieval, 2010, pp. 57-62.

[16] Y. Guo, F. Sohel, M. Bennamoun, J. Wan, M. Lu, A novel local surface feature for 3D object recognition under clutter and occlusion, Information Sciences, 293 (2015) 196-213.

[17] A. Flint, A. Dick, A. van den Hengel, Thrift: Local 3D Structure Recognition, 9th Conference on Digital Image Computing Techniques and Applications, 2007, pp. 182-188.

[18] R.B. Rusu, N. Blodow, Z.C. Marton, M. Beetz, Aligning Point Cloud Views using Persistent Feature Histograms, IEEE/RSJ International Conference on Intelligent Robots and Systems, 2008, pp. 3384-3391.

[19] A. Petrelli, L. Di Stefano, On the repeatability of the local reference frame for partial shape matching, 2011 International Conference on Computer Vision, 2011, pp. 2244 - 2251.

[20] Y. Guo, M. Bennamoun, F.A. Sohel, J. Wan, M. Lu, 3D Free Form Object Recognition using Rotational Projection Statistics, Applications of Computer Vision, IEEE, 2013, pp. 1-8.

[21] R.B. Rusu, N. Blodow, M. Beetz, Fast Point Feature Histograms (FPFH) for 3D Registration, IEEE International Conference on Robotics and Automation, 2009, pp. 3212-3217.

[22] Y. Guo, F. Sohel, M. Bennamoun, M. Lu, J. Wan, Rotational Projection Statistics for 3D Local Surface Description and Object Recognition, International Journal of Computer Vision, 105 (2013) 63-86.

[23] J. Yang, Q. Zhang, Y. Xiao, Z. Cao, TOLDI: An effective and robust approach for 3D local shape description, Pattern Recognition, 65 (2017) 175-187.

[24] S. Salti, F. Tombari, L. Di Stefano, SHOT: Unique signatures of histograms for surface and texture description, Computer Vision and Image Understanding, 125 (2014) 251-264.

[25] A. Mian, M. Bennamoun, R. Owens, On the Repeatability and Quality of Keypoints for Local Feature-based 3D Object Retrieval from Cluttered Scenes, International Journal of Computer Vision, 89 (2010) 348-361.

[26] S. Salti, A. Petrell, F. Tombari, L. Di Stefano, On the Affinity between 3D Detectors and Descriptors, 2012 Second Joint 3DIM/3DPVT Conference: 3D Imaging, Modeling, Processing, Visualization & Transmission, 2012, pp. 424-431.

[27] A. Flint, A. Dick, A. van den Hengel, Local 3D structure recognition in range images, IET Computer Vision, 2 (2008) 208-217.

[28] A. Frome, D. Huber, R. Kolluri, T. Bulow, J. Malik, Recognizing Objects in Range Data Using Regional Point Descriptors, Proceedings of the European Conference on Computer Vision, 2004, pp. 224-237.

[29] Y. Guo, F. Sohel, M. Bennamoun, M. Lu, J. Wan, TriSI: A Distinctive Local Surface Descriptor for 3DModeling and Object Recognition, 8th International Conference on Computer Graphics Theory and Applications (GRAPP), 2013, pp. 86-93.

[30] N.J. Mitra, A. Nguyen, Estimating Surface Normals in Noisy Point Cloud Data, Proceedings of the nineteenth annual symposium on Computational geometry. ACM, 2003, pp. 322-328.

[31] J. Yang, Q. Zhang, Z. Cao, Multi-attribute statistics histograms for accurate and robust pairwise registration of range images, Neurocomputing, 251 (2017) 54-67.

[32] A. Zaharescu, E. Boyer, R. Horaud, Keypoints and Local Descriptors of Scalar Functions on 2D Manifolds, International Journal of Computer Vision, 100 (2012) 78-98.

[33] M. Pauly, Point Primitives for Interactive Modeling and Processing of 3D Geometry, University of Kaiserslautern, 2003.

[34] R.B. Rusu, Semantic 3D object maps for everyday manipulation in human living environments, KI-Künstliche Intelligenz, 2010.

[35] K. Jordan, P. Mordohai, A Quantitative Evaluation of Surface Normal Estimation in Point Clouds, IEEE/RSJ International Conference on Intelligent Robots and Systems, 2014, pp. 4220-4226.

[36] B. Curless, M. Levoy, A Volumetric Method for Building Complex Models from Range Images, 23rd Annual Conference on Computer Graphics and Interactive Techniques, 1996, pp. 303-312.

[37] F. Tombari, S. Salti, L. Di Stefano, Performance Evaluation of 3D Keypoint Detectors, International Journal of Computer Vision, 102 (2013) 198-220.

[38] A.S. Mian, M. Bennamoun, R.A. Owens, A Novel Representation and Feature Matching Algorithm for Automatic Pairwise Registration of Range Images, International Journal of Computer Vision, 66 (2006) 19-40.

[39] B. Taati, M. Greenspan, Local shape descriptor selection for object recognition in range data, Computer Vision and Image Understanding, 115 (2011) 681-694.

[40] P.J. Besl, N.D. McKay, A method for registration of 3-D Shapes, IEEE Transaction on pattern analysis and machine intelligence, 14 (1992) 239-56.

[41] Y. Chen, G. Medioni, Object Modeling by Registration of Multiple Range Images, Image and vision computing, 10 (1992) 145-155.

[42] J.L. Bentley, Multidimensional Binary search trees used for associative searching, Communications of the ACM, (1975) 509-517.

[43] A. Albarelli, E. Rodolà, A. Torsello, Fast and accurate surface alignment through an isometry-enforcing game, Pattern Recognition, 48 (2015) 2209-2226.